# Preprocessing in Inductive Logic Programming

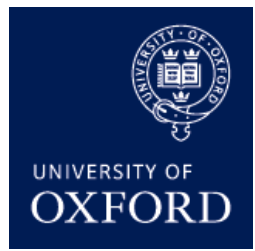

Brad Hunter

Linacre College

University of Oxford

A dissertation submitted for the degree of

*Master of Mathematics and Foundations of Computer Science*

September 15, 2021

Inductive logic programming is a type of machine learning in which logic programs are learned from examples[22]. This learning typically occurs relative to some *background knowledge* provided as a logic program. This dissertation introduces *bottom preprocessing*, a method for generating initial constraints on the programs an ILP system must consider. Bottom preprocessing applies ideas from *inverse entailment* to modern ILP systems. Inverse entailment is an influential early ILP approach introduced with Progol[20]. This dissertation also presents ⊥-Popper, an implementation of bottom preprocessing for the modern ILP system Popper[7]. It is shown experimentally that bottom preprocessing can reduce learning times of ILP systems on hard problems. This reduction can be especially significant when the amount of background knowledge in the problem is large.

# Contents

# List of Figures

# Chapter 1

# Introduction

A common software engineering interview question asks the interviewee to write a program which, given a sequence of characters, determines if the sequence is a palindrome [17]. A palindrome is a sequence of characters that is written the same way forwards and backwards [9]. An interviewer might provide some examples of palindromes. The sequences 'racecar', 'deed', 'mom', 'a', and '' are *positive examples* because they are palindromes. The words 'carriage', 'dead', 'mop', 'at', and 'palindrome' are *negative examples*. If an interviewee struggles, the task can be simplified by providing them a function which reverses a sequence of characters.

Suppose this task is given to a computer instead. The problem of automatically generating a program which satisfies a set of specifications is the domain of program synthesis [10]. This dissertation focuses on Inductive Logic Programming (ILP), a type of machine learning which can be applied to program synthesis. ILP systems learn logic programs from examples. They take as input a set of positive examples, a set of negative examples, and *background knowledge* (BK) in the form of a logic program. In the *learning from entailment* ILP setting used in this dissertation, a logic program *covers* an example if the program logically entails the example relative to the background knowledge. The ideal solution to an ILP problem is a logic program which covers all positive examples and none of the negative examples [19, 22].

Among the advantages of ILP systems are that they can often learn from very few examples [16], they produce human readable and modifiable output, and they support *transfer learning*. Support for transfer learning means that the output of one ILP problem can be used as background knowledge for other problems [29]. This is a desirable property for learning complex systems [6].

**Example 1.1** (Palindrome identification)**.** Palindrome identification can be encoded as an ILP problem using the logic programming language Prolog. The problem can be written as the sets $(E^+, E^-, B)$ where $E^+$ contains the positive examples, $E^-$ contains the negative examples, and $B$ is the background knowledge. We might write the specification given in the first paragraph as follows[1]:

$$E^+ = \left\{ \begin{array}{l} \texttt{palindrome([r, a, c, e, c, a, r]).} \\ \texttt{palindrome([m, o, m]).} \\ \texttt{palindrome([h, e, e, d]).} \\ \texttt{palindrome([a]).} \\ \texttt{palindrome([]).} \end{array} \right\}$$

$$E^- = \left\{ \begin{array}{l} \texttt{palindrome([c, a, r, r, i, a, g, e]).} \\ \texttt{palindrome([m, o, p]).} \\ \texttt{palindrome([p, a, l, i, n, d, r, o, m, e]).} \\ \texttt{palindrome([d, e, a, d]).} \\ \texttt{palindrome([a, t]).} \end{array} \right\}$$

$$B = \left\{ \begin{array}{l} \texttt{last([H], H).} \\ \texttt{last([\_|T],X):- last(T, X).} \\ \texttt{first([H|\_],H).} \\ \texttt{middle([\_|T], T2):- droplast(T, T2).} \\ \texttt{droplast([\_], []).} \\ \texttt{droplast([H|T], [H|T2]):- droplast(T, T2).} \\ \texttt{empty([]).} \\ \texttt{one(1).} \end{array} \right\}$$

In less than a minute, a modern ILP system can find the solution:

```
palindrome(A) :- empty(A).
palindrome(A) :- one(B), length(A,B).
palindrome(A) :- middle(A,C),palindrome(C),last(A,B),first(A,B).
```

Stated a bit informally, this says that a sequence of characters, A, is a palindrome if A is of length zero or one. Otherwise, A is a palindrome if its first and last elements are equal and the rest of the sequence is a palindrome.

Suppose the following definition of reverse is added to $B$ to produce $B'$.

$$B' = B \cup \left\{ \begin{array}{l} \texttt{reverse([], []).} \\ \texttt{reverse([H|T], L):- reverse(T, T2), append(T2,[H],L).} \end{array} \right\}$$

Then a modern ILP system given $B'$ as background knowledge can find the following shorter program almost immediately:

```
palindrome(A) :- reverse(A, A).
```

As with the human interviewee, providing some extra help can simplify the problem tremendously.

---

[1] Lists in Prolog are comma-separated sequences of terms enclosed in square brackets. The syntax [H|T] denotes a relation that holds when H is the head of a list and T is the tail of the list.

As Example 1.1 shows, the choice of background knowledge is critical for ILP problems [4]. Ideally supplying additional background knowledge would reduce learning times. Unfortunately many ILP systems scale poorly as the size of the background knowledge in a problem increases. This sensitivity to the size of the background knowledge limits the practicality of transfer learning in ILP [6].

This dissertation introduces *bottom preprocessing*, a new technique which computes information from the examples of an ILP problem and uses it to reduce the set of programs an ILP system must consider. I show that bottom preprocessing can reduce learning times for an ILP system. Experiments using the Michalski trains problem [13], described in Chapter 5, demonstrate a mean reduction to learning times by over 10x when bottom preprocessing is employed. In a variation of these experiments, I show that bottom preprocessing can also significantly improve the scaling of an ILP system as the background knowledge of a problem increases in size.

This dissertation also presents $\bot$-Popper, an implementation of bottom preprocessing as an extension to the ILP system Popper [7]. Popper encodes an ILP problem as a constraint satisfaction problem (CSP) on the syntax of the logic programming language Prolog. In ILP, possible programs are often called *hypotheses*. The space of possible programs a system explores is called the *hypothesis space*. Popper uses the *learning from failures* technique to generate new constraints whenever it produces a failed hypothesis. These constraints prune hypotheses from its hypothesis space. One of the limitations of Popper is that it only consults examples when testing new hypotheses. This means that its initial hypothesis space is unconstrained by the examples of the problem.

Earlier ILP systems, such as Progol [20] and HAIL [26], use a technique called *inverse entailment*. Rather than search for whole programs as Popper does, they search for clauses one at a time to add to a growing hypothesis. For each clause, they begin by first computing a *bottom clause* from a single positive example. They then use that bottom clause to guide their search for the best clause to add to a hypothesis.

The goal of $\bot$-Popper is to use the bottom clause employed in earlier ILP systems to overcome the initial hypothesis space limitation of Popper. The *bottom clause* of an example $e$ relative to the background knowledge, $\bot_B(e)$, is the logically most specific clause that entails $e$. Informally, the body of $\bot_B(e)$ consists of all facts that cannot be false when $e$ is true under some background knowledge. Any additional facts that could be added to the body of $\bot_B(e)$ are irrelevant or would negate $e$. Thus the bottom clause can be viewed as creating a lower bound on any clause that should appear in a hypothesis.

The key idea of bottom preprocessing is that a hypothesis $H$ can be pruned from the hypothesis space of an ILP problem in the following two cases:

- $e^-$ is a negative example and $H$ logically entails $\perp_B(e^-)$.

- $e^+$ is a positive example and $H$ does not logically entail $\perp_B(e^+)$.

**Example 1.2.** The bottom clause of the negative example `palindrome([a, t])` from Example 1.1, call it $e^-$, relative to the original background knowledge $B$ is:

$$\perp_B(e^-) = \left\{ \begin{array}{l} \text{palindrome(A):-} \\ \quad \text{first(A,B), last(A,C), middle(A,D), empty(D),} \\ \quad \text{length(A,F), length(D,G), one(E).} \end{array} \right\}$$

Suppose $C$ is the clause `palindrome(A):- middle(A,D), empty(D)`. Then $C$ entails $\perp_B(e^-)$. Any hypothesis containing $C$ can be pruned from the hypothesis space of the problem. This makes sense because non-palindromes can have empty middles. Thus $C$ entails too much. Note that, in this case, $C$ is a subset of $\perp_B(e^-)$. Clausal subsets will become important later.

An immediate difficulty arises with this proposed approach because entailment between clauses is undecidable [2]. Thus bottom preprocessing follows inverse entailment approaches in comparing clauses using $\theta$-subsumption [25]. $\theta$-subsumption implies entailment but, while decidable, it is weaker than entailment [24]. As chapter 3 shows, the consequence of this choice is that $\perp$-Popper can, in some cases, prune solutions from Popper's hypothesis space. This means that $\perp$-Popper is incomplete for finding solutions to ILP problems. Experiments in Chapter 5 show that this incompleteness is not a significant limitation on many ILP problems.

$\perp$-Popper is implemented as a wrapper around Popper. It relies on providing hypothesis constraints as input to Popper. A *hypothesis constraint* is a constraint written in the language of Popper's solver which can prune a hypothesis from its hypothesis space. To use hypothesis constraints generated from bottom clauses, they must be written in a form Popper's solver can use. $\perp$-Popper encodes constraints in the Answer Set Programming (ASP) language. It encodes a subsumption check between clauses as well as a set of constraints on hypotheses.

Popper is part of a category of modern ILP systems employing a meta-level approach to ILP. Meta-level ILP systems reason about the structure of logic programs. They often use a solver to search their hypothesis space [5]. While the implementation of $\perp$-Popper is specific to Popper, bottom preprocessing can be applied more broadly to any meta-level ILP system that supports hypothesis constraints.

**Contributions.** This dissertation makes the following contributions:

1. It formalizes the use of bottom preprocessing to generate hypothesis constraints for meta-level ILP systems.

2. It proves that bottom preprocessing of an ILP problem:

   (a) does not prune solutions when applied to negative examples.

   (b) can prune solutions when applied to positive examples. However, there is a well-defined subset of solutions which are never pruned by these constraints.

3. It describes $\bot$-Popper, a working implementation of bottom preprocessing for Popper.

4. It demonstrates empirically that bottom preprocessing can

   (a) reduce learning times on hard ILP problems.

   (b) improve the scaling of meta-level ILP systems as the amount of background knowledge in an ILP problem increases.

**Outline.** This dissertation has the following structure. Chapter 2 establishes the problem setting for bottom preprocessing. Chapter 3 contains theory related to the soundness of bottom preprocessing. Chapter 4 describes the implementation of $\bot$-Popper. Chapter 5 introduces experimental results using the Michalski trains problem [13] and list manipulation programming problems. Chapter 6 discusses related work. Chapter 7 concludes the dissertation with a discussion of the limitations of bottom preprocessing and possible future work.

# Chapter 2

# Problem Setting

This chapter provides[1] an overview of logic programming and then defines the problem setting for bottom preprocessing precisely. The problem setting is an extension of the *learning from failures* (LFF) problem setting described in *Learning programs by learning from failures* [7]. Many of the definitions are only slightly modified from the ones given in this paper.

## 2.1 Logic Programming

Some familiarity with logic programming is assumed. The following brief summary is intended to clearly define the terminology that will be used throughout the dissertation.

### 2.1.1 Syntax

In logic programming, *variables* begin with uppercase letters and *constant symbols* begin with lowercase letters. A *function symbol* is a sequence of lowercase characters. A *term* is a variable, a constant symbol, or a *compound term* of the form $f(t_1, t_2, \cdots, t_n)$ where $f$ is a function symbol and each $t_i$ is a term [8].

A *predicate symbol* is also a sequence of lowercase characters. An *atom* is of the form $p(t_1, t_2, \cdots, t_n)$ where $p$ is a predicate symbol and each $t_i$ is a term. While the syntax of atoms and compound terms are similar, their semantics are different.

The *arity* of a predicate symbol or function symbol defines how many arguments it takes. A predicate symbol (respectively, function symbol) of arity $n$ is usually written $p/n$ ($f/n$) to distinguish predicate symbols (function symbols) of different arities from each other. A comma separated sequence of terms enclosed in parentheses, such as $(t_1, t_2, \cdots, t_n)$, is sometimes referred to as a *tuple*. A comma separated sequence of terms surrounded by square brackets, such as $[t_1, t_2, \cdots, t_n]$, is a *list*. Tuples and lists are compound terms.

---

[1]Error: 'proves'

The symbol $\neg$ indicates negation. A *literal* is either an atom $a$, in which case it is called *positive*, or the negation of an atom $\neg a$, in which case it is called *negative* [24].

A *clause* is a set of literals. A set of clauses is a *clausal theory*. A *Horn clause* is a clause in which no more than one literal is positive. A *definite clause* is a Horn clause with exactly one positive literal. A *logic program* is a clausal theory containing only Horn clauses.[2] A *definite program* or *definite theory* is a clausal theory containing only definite clauses. A term, atom, or clause is *ground* if it contains no variables.

A clause represents the disjunction of its literals. In logic programming, the implication symbol, $\leftarrow$, is written `:-`. Conjunction is represented by commas. Clauses end with full stops. Variables are implicitly universally quantified [5].

The following is a universally quantified Horn clause in first order logic, call it $C$:

$$\forall x, y.\, p(x) \vee \neg q(x, y) \vee \neg r(y)$$

$C$ can be rewritten as an implication:

$$\forall x, y.\, p(x) \leftarrow q(x, y) \wedge r(y)$$

A convention in logic programming is to use uppercase letters at the beginning of the alphabet for variables. Thus, $C$ can be written as the logic program:

```
p(A) :- q(A, B), r(B).
```

The negated literals of a Horn clause are called the *body*. The single positive literal, if it exists, is called the *head*. A clause with no head is a *constraint*. A ground Horn clause with no body is called a *fact*.

A *substitution* of the form $\theta = \{v_1/t_1, v_2/t_2, \cdots, v_n/t_n\}$ is an assignment of terms to variables [8]. The application of a substitution $\theta$ to a clause $C$, written $C\theta$, replaces variables in $C$ with their corresponding terms in $\theta$. If $C$ is the clause `palindrome(A):- middle(A, B), empty(B)` and $\theta$ is the substitution $\{A/[a, t], B/[]\}$ then $C\theta$ is the ground clause `palindrome([a, t]):- middle([a, t], []), empty([])`. Substitutions can also be applied to variables and clausal theories in the same way.

---

[2]Some logic programming languages such as Prolog allow more than just Horn clauses in their syntax. To keep things short, this dissertation uses this simpler definition of a logic program.

### 2.1.2 Semantics

The *vocabulary*, $\mathcal{V}$, of a logic program is the set of all its constant symbols, function symbols and predicate symbols. The *Herbrand universe* is the set of all ground terms which can be constructed from the constant symbols and function symbols in $\mathcal{V}$. The *Herbrand base* is the set of all ground atoms which can be constructed from the predicate symbols in $\mathcal{V}$ and the ground terms in the Herbrand universe. A *Herbrand interpretation* is a subset of the Herbrand base. The atoms in a Herbrand interpretation are treated as true. Any atoms not in the Herbrand interpretation, but in the Herbrand base, are false [5, 24].

**Definition 1** (Herbrand Model). Suppose $C = h \leftarrow b_1, b_2, ..., b_n$ is a clause. A Herbrand interpretation $I$ is a Herbrand model for $C$ if and only if, for all substitutions $\theta$, whenever $h\theta \in i$, all $b_i\theta \in I$ [8].

A Herbrand interpretation $I$ is a Herbrand model of a clausal theory $T$ if $I$ is a Herbrand model of every clause in $T$. A clausal theory $T$ *entails* a clause $C$, written $T \models C$, if every Herbrand model of $T$ is also a Herbrand model of $C$. A clausal theory $T$ is *satisfiable* if $T$ has a Herbrand model and *unsatisfiable* otherwise. A Herbrand interpretation $I$ is a *least Herbrand model* for $T$ if $I$ is a Herbrand model for $T$ and, for all other Herbrand interpretations $I'$ such that $I'$ is a Herbrand model of $T$, $I \subseteq I'$ [21].

If $C$ is a constraint we say that $C$ is *violated* for a clausal theory $T$, if the body of the constraint is in any Herbrand model of $T$. In other words, all of the literals in $C$ cannot be true. If $F$ is a fact in $T$ then $F$ is in every Herbrand model of $T$.

A logic $\mathcal{L}$ is *monotonic* if the set of clauses entailed by a clausal theory $T \in \mathcal{L}$ cannot be reduced by adding a clause $C \in \mathcal{L}$ to $T$. Definite programs are monotonic [8].

### 2.1.3 Logic Programming Languages

**Datalog** is a logic programming language with a subset of the syntax described above. Datalog programs consist of definite clauses. Datalog does not allow nested compound terms. This means that Datalog programs have a finite Herbrand base. Datalog requires that every variable in the head of a clause also appear in the body. These restrictions cause Datalog programs to be decidable and thus guaranteed to terminate[5].

**Prolog** is a popular, Turing complete, logic programming language. It allows the full syntax described in the syntax section. However, Prolog is not a purely declarative language [5]. Despite being very expressive and widely used, Prolog can be quite difficult to reason about precisely. Popper both produces and takes as input Prolog programs.

**Answer Set Programming (ASP)** is a type of logic programming which is convenient for defining and solving constraint satisfaction problems. Its syntax is similar to the logic program syntax described above with extensions not discussed in this dissertation. Popper encodes its CSP problem in ASP.

## 2.2 Problem Definitions

### 2.2.1 Language Bias

The hypothesis space for an ILP system has only been loosely defined thus far. The total set of possible programs an algorithm could potentially explore is infinite in most cases. In Example 1.1 it was implicitly assumed that only predicate symbols which appeared in the background knowledge could be used in a hypothesis. Instead, most ILP systems use a *language bias* as a form of inductive bias to precisely define their hypothesis space [5].

#### 2.2.1.1 Mode Declarations

The inverse entailment setting [20] originated *mode declarations* as a means to establish its language bias. Mode declarations are now used by many ILP systems [20, 27, 26, 15]. The LFF problem setting uses a different but isomorphic set of declarations to define its language bias. Mode declarations are explained briefly here because they are important for the computation of bottom clauses. They are also commonly encountered in ILP literature.

A *mode declaration* has either the form $\text{modeh}(r, atom)$ or $\text{modeb}(r, atom)$. The modeh declarations restrict the form of atoms occurring in the head of a hypothesis clause and the modeb declarations restrict atoms in the body.

**Example 2.1** (Mode declarations). The mode declarations for Example 1.1 might include:

```
modeh(*, palindrome(+list)).
modeb(*, last(+list, -element)).
modeb(*, empty(-list)).
modeb(*, reverse(+list, -list)).
```

Including a modeb declaration for `reverse` gives permission to the ILP system to use the `reverse` predicate symbol in the body of a hypothesis.

The parameter $r$ in a mode declaration is the *recall*. For this dissertation, recall can be ignored. It defaults to $*$ which enforces no bound on hypotheses.

The parameter *atom* is a compound term of the form $p(t_1, t_2, \cdots, t_n)$. The function symbol $p$ in a modeh (respectively modeb) declaration indicates that the predicate symbol $p/n$, with $n$ being the number of arguments in *atom*, can occur in the head (body) of a clause in a hypothesis. The terms $t_i$ are constant symbols made up of a *direction* in $\{+, -, \#\}$ followed by a *type*.

The direction symbols $+, -$ and $\#$ indicate that an argument must be *input*, *output*, or *ground* respectively. *Input* arguments must be variables instantiated in the head of a clause or another literal of the clause before being used. *Output* arguments are variables instantiated by the literal they are used in. *Ground* arguments must be terms [5]. Directions create a strong bias by enforcing a relation between literals in a clause. For instance, the $+$ symbol on the first argument of `modeb(*, last(+list, -element))` means that `palindrome(A):- last(A, B)` is mode consistent but `palindrome(A):- last(C, B)` is not. While directions are required for Prolog, they can be omitted for Datalog.

The *type* enforces a very simple type system. If a variable or term occurs in an atom with a mode declaration that assigns it type $t_1$, then it cannot be assigned a different type $t_2$ by the mode declaration of any other literal in the clause.

### 2.2.1.2 Declaration Definitions

The language bias of the original LFF problem setting does not require recall, directions or types. Directions and types are very useful for bounding the computation of the bottom clause though. The declarations of the LFF problem setting are extended here to include type and direction declarations.

**Definition 2** (Head declaration). A head declaration is a ground atom of the form `head_pred(p,a)` where `p` is a predicate symbol of arity `a`.

**Definition 3** (Body declaration). A body declaration is a ground atom of the form `body_pred(p,a)` where `p` is a predicate symbol of arity `a`.

**Definition 4** (Type declaration). A type declaration is a ground atom of the form `type(p,s)` where `p` is a predicate symbol of arity $a$ and `s` is a tuple of length $a$ containing constant symbols.

**Definition 5** (Direction declaration). A direction declaration is a ground atom of the form `direction(p,s)` where `p` is a predicate symbol of arity $a$ and `s` is a tuple of length $a$ containing only constant symbols in $\{\texttt{in}, \texttt{out}\}$.

These expanded declarations are compatible with the LFF problem setting.[3] Note that these declarations can encode the same information as mode declarations, assuming recall defaults to $*$ and ground arguments are not allowed. Appendix A has a program to convert between the declarations of the expanded LFF problem setting and mode declarations.

**Definition 6** (Language bias). A language bias is a tuple $(L_h, L_b, L_t, L_d)$ with $L_h$ a set of head declarations, $L_b$ a set of body declarations, $L_t$ a set of type declarations, and $L_d$ a set of direction declarations.

**Example 2.2.** Suppose $L = \{L_h, L_b, L_t, L_d\}$ is an example language bias for Example 1.1. If the bias is somewhat reduced to keep the example simple, it might look like the following:

$$L_h = \{\texttt{head\_pred(palindrome,1).}\}$$

$$L_b = \left\{\begin{array}{l}\texttt{body\_pred(first,2).}\\\texttt{body\_pred(last,2).}\\\texttt{body\_pred(middle,2).}\\\texttt{body\_pred(empty,1).}\end{array}\right\}$$

$$L_t = \left\{\begin{array}{l}\texttt{type(palindrome,(list,)).}\\\texttt{type(first,(list,element)).}\\\texttt{type(last,(list,element)).}\\\texttt{type(middle,(list,list)).}\\\texttt{type(empty,(list,)).}\end{array}\right\}$$

$$L_d = \left\{\begin{array}{l}\texttt{direction(palindrome,(in,)).}\\\texttt{direction(first,(in,out)).}\\\texttt{direction(last,(in,out)).}\\\texttt{direction(middle,(in,out)).}\\\texttt{direction(empty,(out,)).}\end{array}\right\}$$

These definitions establish the syntax of the language bias. No indication has been given of the restrictions the declarations actually impose on the hypothesis space. The first step is to define what it means for a clause to be consistent with a given language bias.

---

[3]The Popper implementation supports type and direction declarations though this is not mentioned in the original paper.

**Definition 7** (Language bias consistent clause). Let $L = (L_h, L_b, L_t, L_d)$ be a language bias and $C = h \leftarrow b_1, b_2, \cdots, b_n$ be a definite clause. Then $C$ is language bias consistent with $L$ if and only if:

- $h$ is an atom of the form $p(X_1, \cdots, X_n)$ and $\texttt{head\_pred}(p, n)$ is in $L_h$.

- Every $b_i$ is an atom of the form $p(X_1, \cdots, X_n)$ and $\texttt{body\_pred}(p, n)$ is in $L_b$.

- Every $X_i$ is a variable.

- For all variables $Z$ which occur in $C$, if $Z$ occurs in two literals $l_1 = p(X_1, \cdots, X_n)$ and $l_2 = q(Y_1, \cdots, Y_m)$ as $X_i$ and $Y_j$ respectively, then $\texttt{type}(p, (T_1^p, \cdots, T_n^p))$[4] and $\texttt{type}(q, (T_1^q, \cdots, T_m^q))$ are in $L_t$ and $T_i^p = T_j^q$.

- $a$ is an atom of the form $p(X_1, \cdots, X_n)$ and $\texttt{direction}(p, (D_1^p, \cdots, D_n^p)) \in L_d$ then:

    - If $a = h$ and $D_i^p$ is $\texttt{out}$, then one of the following must be true:
        * the variable $X_i$ must equal some variable $X_j$ in $h$ and $D_j^p$ must be $\texttt{in}$.
        * the variable $X_i$ must occur in some $b_k = q(Y_1, \cdots, Y_m)$ as some $Y_j$ and $D_j^q$ must be $\texttt{out}$ in $\texttt{direction}(q, (D_1^q, \cdots, D_m^q)) \in L_d$.
    
    - If $a = b_j$ and $D_i^p$ is $\texttt{in}$, then the variable $X_i$ must occur in some $c = q(Y_1, \cdots, Y_m)$ as $Y_j$. There must also be $\texttt{direction}(q, (D_1^q, \cdots, D_m^q)) \in L_d$ and either $c = h$ and $D_j^q$ is $\texttt{in}$ or $c = b_x$ and $D_j^q$ is $\texttt{out}$.

We also extend Popper's definition of a declaration bias consistent hypothesis to cover type and direction declarations.

**Definition 8** (Language bias consistent hypothesis). A language bias consistent hypothesis $H$ relative to a language bias $L$ is a definite clausal theory where each clause $C \in H$ is language bias consistent with $L$.

**Example 2.3** (Language bias consistent hypotheses). Given the language bias $L$ from Example 2.2, the following hypotheses are language bias consistent with $L$:

- `palindrome(A):- last(A, B), middle(A, C), empty(C).`

- `palindrome(A):- first(A, B), last(A, B).`

---

[4] Error: $S_1^p$

- `palindrome(A):- empty(A).`
  `palindrome(A):- first(A, C), middle(A, B), last(A, C).`

**Example 2.4** (Language bias inconsistent hypotheses). Given the language bias $L$ from Example 2.2, the following hypotheses are **not** language bias consistent with $L$:

- `palindrome(A):- length(A, B), one(B).`
  `length/2` and `one/1` are not allowed as predicate symbols in the body of the clause because `body_pred(length, 2)` and `body_pred(one, 1)` are not in $L_b$.

- `empty(A):- middle(A, C), empty(C).`
  `empty/1` is not an allowed predicate symbol in the head of the clause because `head_pred(empty, 1)` is not in $L_h$.

- `palindrome(A):- first(A), empty(A).`
  `first` cannot appear with arity 1 because `body_pred(first, 1)` is not in $L_b$.

- `palindrome(A):- first(A, B), empty(B).`
  The types do not match. `B` must have type `element` from `type(first, ( list, element))` in $L_t$. However, `type(empty, (list))` in $L_t$ requires $B$ to have type `list`[5].

- `palindrome(A):- last(B, A), first(B, A).`
  `palindrome(A):- middle(A, B), empty(B).`
  The direction of `B` in the first clause is incorrect. Due to `direction(first ,(in,out))` and `direction(last,(in,out))` in $L_d$, `B` is an input variable. `B` does not occur as an output variable in another literal in the body or as an input variable in the head.

### 2.2.2 Hypothesis constraints

Bottom preprocessing operates by generating a set of hypothesis constraints for a meta-level ILP system. The specific language of these constraints can differ depending on the target ILP system. Thus, to keep the problem setting general, hypothesis constraints are defined in terms of how they affect the hypothesis space. The specific hypothesis language used by the $\bot$-Popper implementation is given in Chapter 4.

---
[5]Error: element

**Definition 9** (Hypothesis constraint). Let $\mathcal{L}^6$ be a language that defines hypotheses, i.e. a meta-language. Then a *hypothesis constraint* is a constraint expressed in $\mathcal{L}$.

**Example 2.5.** Suppose $H$ is the hypothesis from Example 1.2:

```
palindrome(A) :- middle(A, B), empty(B).
```

Suppose the hypothesis language $\mathcal{L}$ is ASP with an encoding of hypotheses similar to the one used by Popper. An explanation of the specific encoding Popper uses is described in Chapter 4. For now it suffices to know that $H$ could be encoded in Popper as:

```
clause(0).
head_literal(0, palindrome, 1, (0,)).
body_literal(0, middle, 2, (0, 1)).
body_literal(0, empty, 1, (1,)).
```

Suppose `clause_size(C, N)` holds when the clause $C$ contains exactly $N$ literals. Then, to banish any hypothesis containing exactly the single clause in $H$, a hypothesis constraint, call it $C_1$ could be written as follows:

```
:- clause(C),
   head_literal(C, palindrome, 1, (0,)),
   body_literal(C, middle, 2, (0, 1)),
   body_literal(C, empty, 1, (1,)),
   clause_size(C, 3).
```

**Definition 10** (Hypothesis Space). Suppose $B$ is a language bias and $C$ is a set of hypothesis constraints in $\mathcal{L}$. Then the *hypothesis space*, $\mathcal{H}_{B,C}$, is the set of all bias consistent, relative to $B$, hypotheses which, when written in $\mathcal{L}$, do not violate any constraints in $C$.

**Example 2.6.** Suppose $B$ is a language bias and $\mathcal{H}_B$ is the set of all hypotheses written in $\mathcal{L}$ that are language bias consistent with $B$. Suppose $H_1$, $H_2$ and $H_3$ are hypotheses consistent with $B$ defined as follows:

$$H_1 = \{\texttt{palindrome(A):- middle(A, B), empty(B).}\}$$
$$H_2 = \left\{ \begin{array}{l} \texttt{palindrome(A):- middle(A, B), empty(B).} \\ \texttt{palindrome(A):- length(A, B), one(B).} \end{array} \right\}$$
$$H_3 = \{\texttt{palindrome(A):- length(A, B), one(B).}\}$$

Suppose $C = \{C_1\}$ is a set of hypothesis constraints written in $\mathcal{L}$ containing the constraint $C_1$ from Example 2.5. Then $\{H_1, H_2, H_3\} \subseteq \mathcal{H}_B$ but $H_1 \notin \mathcal{H}_{B,C}$ and $H_2 \notin \mathcal{H}_{B,C}$.

---

[6] Several errors used $\mathcal{L}_\mathcal{H}$ here

Suppose $B$ is a language bias. Suppose $C$ and $C'$ are sets of hypothesis constraints. Then the hypothesis space generated by the combined set $C \cup C'$ is written $\mathcal{H}_{B,C\cup C'}$. Suppose $H$ is a hypothesis. If $H \in \mathcal{H}_{B,C\cup C'}$ then $H$ does not violate any constraint in $C \cup C'$. Thus $H \in \mathcal{H}_{B,C}$ and $H \in \mathcal{H}_{B,C'}$. Similarly, it follows that if $H \in \mathcal{H}_{B,C}$ and $H \in \mathcal{H}_{B,C'}$ then $H \in \mathcal{H}_{B,C\cup C'}$. Thus $\mathcal{H}_{B,C\cup C'} = \mathcal{H}_{B,C} \cap \mathcal{H}_{B,C'}$.

### 2.2.3 Problem Input

Having defined the language bias and hypothesis constraints of an LFF problem, it is now possible to give a full description of an LFF problem input.

**Definition 11** (LFF Problem Input). A *problem input* is a tuple $(B, L, C, E^+, E^-)$ where

- $B$ is a Horn program denoting background knowledge. $B$ has the following restrictions:
    - For all $e \in E^+ \cup E^-$, it must be the case that $B \not\models e$.
    - If $p/a$ is a predicate symbol such that `head_pred(p, a)` $\in L_h$ then $p/a$ cannot appear in the body of any clause in $B$.

- $L$ is a language bias.

- $H$ is a set of hypothesis constraints.

- $E^+$ is a set of facts denoting positive examples.

- $E^-$ is a set of facts denoting negative examples.

The restrictions on the BK exist to ensure three things:

- A problem is not trivially unsatisfiable because the BK entails a negative example.

- No positive example is irrelevant, as it would be if the BK entailed the example without a hypothesis.

- A clause in the BK cannot resolve with a clause in the hypothesis. This ensure a hypothesis is not made recursive due to a clause in the BK.

### 2.2.4 Solutions

With LFF problem input defined, it is now possible to give a formal definition of what it means for a hypothesis to be a solution to a problem in the LFF setting.

**Definition 12** (Solution). Given an input tuple $(B, L, C, E^+, E^-)$, a hypothesis $H \in \mathcal{H}_{L,C}$ is a *solution* when

- $\forall e \in E^+.\ B \cup H \models e$. The solution is *complete*.

- $\forall e \in E^-.\ B \cup H \not\models e$. The solution is *consistent*.

The ability to handle misclassified examples, a type of *noise*, is important in many types of machine learning [8]. The LFF problem setting, and by extension bottom preprocessing, assume that all examples are correctly classified. If noise exists in the examples, an LFF system might be unable to find a solution.

Many meta-level ILP systems will learn an *optimal solution* if one exists [5]. There are potentially many ways to define optimality. In the LFF problem setting, optimality is defined in terms of program size.

**Definition 13** (Optimal Solution). Let $P = (B, L, C, E^+, E^-)$ be a problem input. Let $H \in \mathcal{H}_{L,C}$ be a hypothesis. Then $H$ is an *optimal solution* if both:

1. $H$ is a solution to $P$.

2. for all $H' \in \mathcal{H}_{L,C}$ where $H'$ is a solution, the number of literals in $H'$ is greater than or equal to the number of literals in $H$.

## 2.3 Generalization and Specialization

Deciding entailment between clauses is undecidable [2]. Due to this, inverse entailment based systems test entailment between clauses using a weaker relation called subsumption.

**Definition 14** (Clausal Subsumption). A clause $C_1$ *subsumes* a clause $C_2$ if and only if there exists a substitution $\theta$ such that $C_1\theta \subseteq C_2$.

**Example 2.7** (Subsumption). Suppose the clauses $C_1$, $C_2$, and $C_3$ are defined as follows:

```
C₁ = palindrome(A):- first(A, B), middle(A, C), last(A, B).
C₂ = palindrome(A):- first(A, C), middle(A, B).
C₃ = palindrome(A):- first(A, B), last(A, B).
```

Then $C_3 \subseteq C_1$ so $C_3$ subsumes $C_1$. $C_2\theta \subseteq C_1$ with substitution $\theta = \{C/B, B/C\}$ so $C_2$ subsumes $C_1$. $C_1$ does not subsume either $C_2$ or $C_3$. Finally, $C_2$ does not subsume $C_3$ and $C_3$ does not subsume $C_2$.

If a clause $C_1$ subsumes $C_2$ then $C_1 \models C_2$. However, the converse is not necessarily true. We cannot conclude that if $C_1 \models C_2$ then $C_1$ subsumes $C_2$ [24].

Since most meta-level ILP systems search for entire clausal theories [6], it's useful to extend subsumption to clausal theories. The theory that follows this chapter will also operate mostly at the level of clausal theories rather than clauses.

**Definition 15** (Theory subsumption). A Horn theory $T_1$ subsumes a Horn theory $T_2$, denoted $T1 \preceq T2$, if and only if $\forall C_2 \in T_2, \exists C_1 \in T_1$ such that $C_1$ subsumes $C_2$.

If $T_1$ and $T_2$ are clausal theories and $T_1 \preceq T_2$ then $T_1 \models T_2$ [7]. Again, the converse is not necessarily true.

It is worth noting that a single clause can also be treated like a one element clausal theory. We sometimes say that a clausal theory $T$ is a generalization/specialization of a clause $C$. When we do this, we are implicitly comparing $T$ to the clausal theory $T_C$ which is the one element set consisting only of $C$.

Rather than using subsumption directly, it can be useful to consider generalizations and specializations of a clausal theory.

**Definition 16** (Generalization). A clausal theory $T_1$ is a generalization of a clausal theory $T_2$ if and only if $T_1 \preceq T_2$.

**Definition 17** (Specialization). A clausal theory $T_1$ is a specialization of a clausal theory $T_2$ if and only if $T_2 \preceq T_1$.

Finally, it will be useful to define a *most specific generalization*. Note that a theory which is a most specific generalization of another theory is not necessarily unique.

**Definition 18** (Most specific generalization). A clausal theory $T_1$ is a *most specific generalization* of another clausal theory $T_2$ if $T_1 \preceq T_2$ and for all other clausal theories $T_3$ such that $T_3 \preceq T_2$, $T_3 \preceq T_1$.

# Chapter 3

# Preprocessing Theory

The fundamental idea of this dissertation is to use information about the examples and background knowledge in an ILP problem to prune its initial hypothesis space. This can be done by preprocessing the examples to generate a set of additional hypothesis constraints.

**Remark.** *Throughout this chapter, the BK of an LFF problem input is assumed to be a definite program with a finite Herbrand base.*

A consequence of this remark is that the BK is monotonic and does not allow function symbols, also called *function-free*. In Chapter 4, methods are described for handling infinite Herbrand bases in practice.

## 3.1 Preprocessing Problem

**Definition 19** (Preprocessing problem). Suppose $P = (B, L, C, E^+, E^-)$ is an LFF problem input. Suppose $|\mathcal{H}_{L,C}| = n$ is the number of hypotheses in the hypothesis space of $P$. The *preprocessing problem* is to generate a set of hypothesis constraints $C'$ such that $|\mathcal{H}_{L,C\cup C'}| < n$.

Note that the hypothesis space of an LFF problem input is bounded in practice and thus finite for a given language bias $L$ [7]. This means it is possible to compute both $|\mathcal{H}_{L,C}|$ and $|\mathcal{H}_{L,C\cup C'}|$. A description of this bound is given in Subsection 3.2.4.

Ideally the preprocessing problem can be solved without pruning any optimal solutions from the hypothesis space. In fact, the best possible solution to the preprocessing problem would also prune any hypothesis that is not an optimal solution from the hypothesis space.

**Definition 20** (Optimal preprocessing)**.** Suppose $P = (B, L, C, E^+, E^-)$ is an LFF problem input. Suppose $H$ is a hypothesis in $\mathcal{H}_{L,C}$. The *optimal preprocessing* problem is to generate a set of hypothesis constraints $C'$ such that $H \in \mathcal{H}_{L,C \cup C'}$ if and only if $H$ is an optimal solution to $P$.

Suppose $C'$ is a set of hypothesis constraints generated by optimal preprocessing of some LFF problem input $P = (B, L, C, E^+, E^-)$. It is clear from Definition 20 that any system which can compute $\mathcal{H}_{L,C \cup C'}$ can trivially solve $P$ optimally by selecting any $H \in \mathcal{H}_{L,C \cup C'}$.

In other words, optimal preprocessing is a solution to the general ILP problem. There are many classes of ILP problems which can be encoded as LFF problem input and are not polynomial time learnable [24].

Perhaps a better approach than optimal preprocessing is to consider cases when preprocessing is guaranteed not to prune any optimal solutions from a hypothesis space. First it can be useful to define soundness for hypothesis constraints.

**Definition 21** (Sound hypothesis constraints)**.** Suppose $P = (B, L, C, E^+, E^-)$ is an LFF problem input. Then a set of hypothesis constraints $C'$ is *sound* for $P$ if, for all $H \in \mathcal{H}_{L,C}$ such that $H$ is an optimal solution to $P$, $H \in \mathcal{H}_{L,C \cup C'}$.

Sound preprocessing generates sound hypothesis constraints.

**Definition 22** (Sound preprocessing)**.** Suppose $P = (B, L, C, E^+, E^-)$ is an LFF problem input. The *sound preprocessing* problem is to generate a set of hypothesis constraints $C'$ such that $C'$ is sound for $P$.

## 3.2 Bottom Preprocessing

Preprocessing is an option for any meta-level ILP system that supports hypothesis constraints. There are potentially many different approaches to hypothesis constraints that could be generated with preprocessing. The remainder of this dissertation focuses on hypothesis constraints generated from the bottom clauses of the examples in an ILP problem. This approach, called *bottom preprocessing*, is defined formally at the end of this section.

### 3.2.1 Bottom Clauses

**Definition 23** (Bottom clause). Given background knowledge $B$ and a fact $e$, the *bottom clause* of $e$ relative to $B$, $\perp_B(e)$, is the set of all literals $\neg L$ such that $B \cup \neg e \models L$.

Bottom clauses can be computed for clauses. However, the LFF problem setting restricts examples to be facts. Since this dissertation is only concerned with the bottom clauses of examples, this definition is restricted to the bottom clauses of facts. Suppose $B$ is background knowledge, $C$ is a clause and $e$ is a fact. The motivation for this definition of the bottom clause is that if $B \cup C \models e$ then $B \cup \neg e \models \neg C$ [11].

The bottom clause of a fact $e$ given background knowledge $B$ always exists and is finite as long as the following conditions are met [21]:

- The Herbrand base of $B$ is finite.

- $B \cup \neg e$ is satisfiable. i.e. $B$ is consistent and $B \not\models e$.

Bottom clauses were introduced as part of the *inverse entailment* approach to ILP [20]. Inverse entailment underlies the Progol and Aleph systems as well as many successors [27, 26, 31].

**Definition 24.** A clause $C$ can be derived by *inverse entailment* from an example $e$ given background knowledge $B$ if and only if $C$ subsumes $\perp_B(e)$.

Yamamoto noted that inverse entailment might more accurately be called inverse subsumption to better distinguish it from other approaches that do not use subsumption to invert entailment [31]. This dissertation uses inverse entailment in its original form as given by Definition 24.

According to Definition 23, the bottom clause for the negative example `palindrome([a, t])` from Example 1.1 [1] is:

```
palindrome([a, t]) :- first([a, t],a), last([a, t],t), middle([a,
   t],[]), empty([]), length([a, t],2), length([a, t],0), one(1).
```

This is a more specific clause than the bottom clause given in Example 1.2. That is because this clause is ground and thus not bias consistent. A bottom clause can be *lifted* to be bias consistent by replacing constants with variables and removing literals which violate bias declarations. This process results in a generalization of the bottom clause. The definition below uses the equivalent formulation of the bottom clause $\perp(e)$ as the most specific clause such that $B \cup \perp_B(e) \models e$ [28].

---

[1]Note that this running example is not function-free. It is used here purely for illustrative purposes. Chapter 4 discusses how BK that is not function-free is handled in practice.

**Definition 25** (Bias consistent bottom clause). Given an LFF problem input $P = (B, L, C, E^+, E^-)$, the *bias consistent bottom clause* of an example $e \in E^+ \cup E^-$ relative to $B$ and $L$, denoted $\perp_{B,L}(e)$, is the most specific bias consistent clause such that $B \cup \perp_{B,L}(e) \models e$.

Note that the bias consistent bottom clause always exists if the bottom clause does. In the extreme case, the bias consistent bottom clause is the example fact lifted to variables. Chapter 4 describes a bottom clause algorithm which finds the bias consistent bottom clause of an example directly.

The next proposition is the backbone of the bottom preprocessing approach. It shows that a hypothesis that generalizes the bottom clause of an example entails, relative to the background knowledge, the example itself. This is a well known result in inverse entailment and could be presented as a corollary of the original presentation [20]. It is re-derived here to align it with the other definitions and expand the original result to definite clausal theories.

**Proposition 1.** *Given an LFF problem input $P = (B, L, C, E^+, E^-)$ and a definite clausal theory $T$, if $e \in E^+ \cup E^-$ and $T$ is a generalization of $\perp_{B,L}(e)$ then $B \cup T \models e$.*

*Proof.* Suppose $P$ and $T$ are as stated, $e \in E^+ \cup E^-$ and $T$ is a generalization of $\perp_{B,L}(e)$. Then $T \models \perp_{B,L}(e)$ by the definition of generalization and the fact that subsumption implies entailment. $\perp_{B,L}(e)$ is a definite clause since it is bias consistent. $B$ and $T$ are also definite programs by definition. Since definite programs are monotonic, $B \cup T \models B \cup \perp_{B,L}(e)$. $B \cup \perp_{B,L}(e) \models e$ by the definition of the bias consistent bottom clause so, by transitivity of entailment, $B \cup T \models e$. □

### 3.2.2 Negative Bottom Preprocessing

Bottom preprocessing generates two different sets of hypothesis constraints. One set is generated by preprocessing the bottom clauses of negative examples in an LFF problem. The other set is generated by preprocessing the bottom clauses of positive examples. This subsection defines negative bottom preprocessing as the approach which generates hypothesis constraints from negative examples.

**Definition 26** (Negative bottom preprocessing). Suppose $P = (B, L, C, E^+, E^-)$ is an LFF problem input. Suppose $H \in \mathcal{H}_{L,C}$. *Negative bottom preprocessing* generates a set of hypothesis constraints $C'$ such that $H \in \mathcal{H}_{L,C \cup C'}$ if and only if, for all examples $e^- \in E^-$, $H$ is not a generalization of $\perp_{B,L}(e^-)$.

Using Proposition 1, it is straightforward to show that the hypothesis constraints generated by negative bottom preprocessing are sound.

**Proposition 2.** *Negative bottom preprocessing is sound.*

*Proof.* Suppose $P = (B, L, C, E^+, E^-)$ is an LFF problem input. Suppose $H \in \mathcal{H}_{L,C}$. Suppose $C'$ is a set of hypothesis constraints generated by negative bottom preprocessing of $P$. To show negative bottom preprocessing is sound, it suffices to show $C'$ is a sound set of hypothesis constraints. To show $C'$ is a sound set of hypothesis constraints, it is necessary to show that if $H \notin \mathcal{H}_{L,C \cup C'}$ then $H$ is not an optimal solution to $P$.

Suppose $H \notin \mathcal{H}_{L,C \cup C'}$. Since $C'$ was generated by negative bottom preprocessing, it must be the case that $H$ is a generalization of $\bot_{B,L}(e^-)$ for some $e^- \in E^-$. Any hypothesis in $\mathcal{H}_{L,C}$ is a definite clausal theory by definition. By Proposition 1 $B \cup H \models e^-$. Then, by Definition 12, $H$ is not consistent and cannot be a solution to $P$. □

### 3.2.3 Positive Bottom Preprocessing

This subsection defines *positive bottom preprocessing*, the portion of the bottom preprocessing approach which generates hypothesis constraints from positive examples of an ILP problem.

**Definition 27** (Positive bottom preprocessing)**.** Suppose $P = (B, L, C, E^+, E^-)$ is an LFF problem input. Suppose $H \in \mathcal{H}_{L,C}$. *Positive bottom preprocessing* generates a set of hypothesis constraints $C'$ such that $H \in \mathcal{H}_{L,C \cup C'}$ if and only if, for all examples $e^+ \in E^+$, $H$ is a generalization of $\bot_{B,L}(e^+)$.

It would be ideal if positive bottom preprocessing were provably sound. However, a hypothesis which is **not** a generalization of the bottom clause of a positive example may still be a solution to an ILP problem. As noted when subsumption was defined, for clausal theories $T_1$ and $T_2$, if $T_1 \models T_2$ it may not be the case that $T_1 \preceq T_2$. The contrapositive then also does not hold. If $T_1 \npreceq T_2$ it may still be the case that $T_1 \models T_2$.

The classic example of the incompleteness of subsumption for inverse entailment given by Yamamoto [30] was updated to be function-free by Muggleton [21]. It is as follows:

**Example 3.1.** Suppose $B$ is background knowledge, $e^+$ is a positive example, $\bot_B(e^+)$ is the bottom clause of $e^+$, and $H$ is a hypothesis. Suppose they have the form:

$$B = \left\{ \begin{array}{l} \texttt{even(0).} \\ \texttt{zero(0).} \\ \texttt{even(A):- succ(B, A), odd(B).} \end{array} \right\}$$

$$e^+ = \{\texttt{odd(3):- succ(0, 1), succ(1, 2), succ(2, 3).}\}$$

$$H = \{\texttt{odd(A):- succ(B, A), even(B)}\}$$

$$\bot_B(e^+) = \{\texttt{odd(A):-zero(B),even(B),succ(B, C),succ(C, D),succ(D, A).}\}$$

Then $H$ is not a generalization of $\bot_B(e^+)$ but $B \cup H \models e^+$.

Other approaches to characterizing bottom were devised which could find hypotheses outside the subsumption lattice between the bottom clause and the empty clause [26]. This dissertation uses inverse entailment and the bottom clause as given in the original paper [20] and acknowledges their incompleteness. In future work, other techniques to generate hypothesis constraints for positive examples could be attempted using improvements on inverse entailment.

The incompleteness of inverse entailment means that positive bottom preprocessing is unsound. It is possible, however, to show that positive bottom preprocessing does not prune a specific subset of hypotheses defined by relative subsumption [25].

**Definition 28** (Relative subsumption). A clause $C_1$ subsumes a clause $C_2$ relative to a clausal theory $B$, written $C_1 \succeq_B C_2$ if there is a substitution $\theta$ such that $B \models \forall (C_1 \theta \to C_2)$ [24].

The $\forall$ symbol in this definition quantifies over all possible assignments of ground terms in the Herbrand universe to the variables in $C_1\theta$ and $C_2$. An alternative definition of relative subsumption characterizes it as follows. Suppose $T = B \cup C_1$ is a definite theory. Suppose $T \vdash_c C_2$ if and only if $C_2$ can be derived by resolution from $T$ using each clause in $T$ a maximum of one time. Then $C_1 \succeq_B C_2$ if and only if $T \vdash_c D$ for some clause $D$ such that $D$ subsumes $C_2$ [22].

**Example 3.2** (Relative Subsumption). This example of relative subsumption is given in *Foundations of Inductive Logic Programming* [24]. Suppose the following:

$$C_1 = \{\texttt{small(X):- cat(X).}\}$$

$$C_2 = \{\texttt{cuddly\_pet(X):- fluffy(X), cat(X).}\}$$

$$B = \left\{ \begin{array}{l} \texttt{pet(X):- cat(X).} \\ \texttt{cuddly\_pet(X):- small(X), fluffy(X), pet(X).} \end{array} \right\}$$

Then $C_1 \succeq_B C_2$.

For Horn theories $T_1$, $T_2$ and $B$, $T_1 \succeq_B T_2$ if, for all clauses $C_2 \in T_2$, there exists a clause $C_1 \in T_1$ such that $C_1 \succeq_B C_2$. Note that if $T_1 \preceq T_2$ then it is also the case that $T_1 \succeq_B T_2$ for any Horn theory $B$. Further, if $T_1 \succeq_B T_2$ then $B \cup T_1 \models T_2$. The inverse is not true for either case. This means subsumption is weaker than relative subsumption which is weaker than entailment [24].

**Definition 29** (Subsumption-complete solution). Suppose $P = (B, L, C, E^+, E^-)$ is an LFF problem input. Suppose $H$ is a solution to $P$. $H$ is a *subsumption complete solution* if for all $e^+ \in E^+$ it is the case that $H \succeq_B e^+$.

Suppose $P = (B, L, C, E^+, E^-)$ is an LFF problem input. Suppose $H \in \mathcal{H}_{L,C}$ is a subsumption-complete solution to $P$. Then for all $e^+ \in E^+$, $H \cup B \models e^+$. Thus it is correct to define the set of subsumption-complete solutions to $P$ as a subset of the set of solutions to $P$. However, since for some $e^+ \in E^+$, if $H \cup B \models e^+$ it may not be the case that $H \succeq_B e^+$, the set of solutions and set of subsumption-complete solutions are not equal.

**Definition 30** (Subsumption sound preprocessing). Suppose $P = (B, L, C, E^+, E^-)$ is an LFF problem input. Suppose $H \in \mathcal{H}_{L,C}$. Then *subsumption-sound preprocessing* generates a set of hypothesis constraints $C'$ such that if $H$ is a subsumption-complete solution to $P$, $H \in \mathcal{H}_{L, C \cup C'}$.

With these definitions it is possible to show that positive bottom preprocessing is subsumption-sound. This proof relies on Theorem 5 from Yamamoto [30] which shows that inverse entailment is complete with respect to relative subsumption if the example is not a tautology and the background knowledge does not entail the example.

**Proposition 3.** *Positive bottom preprocessing is subsumption-sound.*

*Proof.* Suppose $P = (B, L, C, E^+, E^-)$ is an LFF problem input, $H \in \mathcal{H}_{L,C}$, and $e^+ \in E^+$. Suppose $C'$ is generated by positive bottom preprocessing of $P$. Suppose $H \notin \mathcal{H}_{L, C \cup C'}$. To show positive bottom preprocessing is sound, it suffices to show that $H$ is not a subsumption-complete solution to $P$.

Suppose, for contradiction, that $H$ is a subsumption complete solution to $P$. There must be some clause $D \in H$ such that $D \succeq_B e+$. Since $e^+$ is not a tautology and $B \not\models e^+$ by the definition of an LFF problem, Yamamoto's theorem applies. This means that $D$ can be found by inverse entailment. It must be the case that $D$ subsumes $\perp_B(e^+)$. This means $B \cup D \models e^+$. $D$ must be bias consistent if it can be found by inverse entailment. From the definition of the bias consistent bottom clause, it must be the case that $D$ subsumes $\bot_{B,L}(e^+)$. But then since $e+$ is arbitrary, $H$ is a generalization of $\bot_{B,L}(e^+)$ for all $e^+ \in E^+$. This contradicts the assumption that $H \notin \mathcal{H}_{L,C \cup C'}$. □

Thus positive bottom preprocessing does not prune a well-defined subset of solutions to an ILP problem. However, there is no guarantee that a subsumption-complete solution to a problem exists. Thus positive bottom preprocessing may prune all solutions and is not guaranteed to preserve optimality. The degree to which this is a problem for positive bottom preprocessing is explored experimentally in Chapter 5.

### 3.2.4 Bottom Preprocessing Definition

Having defined positive and negative preprocessing, it is possible to give a full definition of *bottom preprocessing*.

**Definition 31** (Bottom preprocessing). Suppose $P = (B, L, C, E^+, E^-)$ is an LFF problem input. Suppose $H \in \mathcal{H}_{L,C}$. *Bottom preprocessing* generates a set of hypothesis constraints $C'$ such that $H \in \mathcal{H}_{L,C \cup C'}$ if and only if $C' = C^+ \cup C^-$ where all of the following hold:

- $C^-$ is generated by negative bottom preprocessing of $P$.
- $C^+$ is generated by positive bottom preprocessing of $P$.
- $H \in \mathcal{H}_{L,C \cup C^+}$ and $H \in \mathcal{H}_{L,C \cup C^-}$.

To determine the number of hypotheses pruned by bottom preprocessing, suppose that $P = (B, L, C, E^+, E^-)$ is and LFF problem input and its initial hypothesis space is unconstrained. Thus $C = \{\}$. The set of hypotheses pruned by bottom preprocessing is then described by:

$$\mathcal{H}_{L,\{\}} \setminus (\mathcal{H}_{L,C^+} \cap \mathcal{H}_{L,C^-})$$

In the original LFF paper [7], the unconstrained hypothesis size for an LFF problem was given an upper bound. Suppose $L$ is a language bias with maximum arity $a$ for any predicate symbol in $L_b$ or $L_h$. Suppose $v$ is the maximum number of variables allowed in a clause, $m$ is the maximum number of body literals allowed in a clause and $n$ is the maximum number of clauses allowed in a hypothesis. Then the maximum number of hypotheses in $\mathcal{H}_{L,\{\}}$ was proven to be:

$$\sum_{j=1}^{n} \binom{|L_h|v^a \sum_{i=1}^{m} \binom{|L_b|v^a}{i}}{j}$$

The expression $|L_h|v^a \sum_{i=1}^{m} \binom{|L_b|v^a}{i}$ defines the size of the set of all clauses that can exist in a hypothesis in $\mathcal{H}_{L,\{\}}$. Call that set of clauses $c$. Define $c^+$ to be the subset of clauses in $c$ that generalize $\perp_{B,L}(e^+)$ for at least one $e^+ \in E^+$. Then define $c^-$ to be the subset of clauses in $c$ that generalize $\perp_{B,L}(e^-)$ for at least one $e^- \in E^-$. Then $|L_h|v^a \sum_{i=1}^{m} \binom{|L_b|v^a}{i} - |c^+| + |c^+ \cap c^-|$ is the total number of clauses that either do not generalize a positive example bottom clause or generalize a negative example bottom clause.

Finally, suppose $h$ is the set of all hypotheses $H$ in which all clauses in $H$ are in $c^+$ but $H$ does not generalize $\perp_{B,L}(e^+)$ for all $e^+ \in E^+$. Then the total number of hypotheses pruned by bottom preprocessing can be characterized, very loosely, as follows:

$$\sum_{j=1}^{n} \binom{|L_h|v^a \sum_{i=1}^{m} \binom{|L_b|v^a}{i} - |c^+| + |c^+ \cap c^-|}{j} + |h|$$

In the worst case, all hypotheses generalize the bottom clauses of all positive examples ($|h| = 0$ and $|c^+| = |L_h|v^a \sum_{i=1}^{m} \binom{|L_b|v^a}{i}$) and no possible clauses generalize the bottom clause of a negative example ($|c^-| = 0$). As long as that is not the case, it is safe to assume that bottom preprocessing will prune some hypotheses from $\mathcal{H}_{L,\{\}}$.

## 3.3 Generalization Relations

Underlying bottom preprocessing is a generalization relation based on subsumption. Subsumption testing between clauses is NP-complete [8]. Chapter 5 shows experimentally that the full subsumption check can sometimes significantly increase learning times for hard ILP problems. The remainder of this chapter proves ways in which the generalization relation used by bottom preprocessing can be weakened without pruning any extra solutions.

Chapter 2 used the symbol $\preceq$ to represent subsumption. The generalization relations discussed here use the symbols $\preceq_s$ and $\preceq_c$ to indicate that they are a weakened form of subsumption. Definitions of soundness and completeness are once again given, this time for generalization relations.

**Definition 32** (Subsumption-sound generalization relation). A generalization relation $\preceq_s$ is *subsumption-sound* if for all clausal theories $A, B$ such that $A \preceq_s B$, $A \preceq B$.

**Definition 33** (Subsumption-complete generalization relation). A generalization relation $\preceq_c$ is *subsumption-complete* if for all clausal theories $A, B$ such that $A \preceq B$, $A \preceq_c B$.

### 3.3.1 Negative Bottom Preprocessing

It is now possible to extend Proposition 2 to show that negative bottom preprocessing is still sound if a subsumption-sound generalization relation $\preceq_s$ is used.

**Proposition 4.** *Suppose $P = (B, L, C, E^+, E^-)$ is an LFF problem input. Suppose $\preceq_s$ is a subsumption-sound generalization relation. Suppose $C'$ is a set of hypothesis constraints generated by negative bottom preprocessing with $\preceq_s$ defining generalization. Then $C'$ is sound.*

*Proof.* Suppose $P$, $\preceq_s$, and $C'$ are as stated. Suppose $e^- \in E^-$ and $H \in \mathcal{H}_{L,C}$.

Suppose $H \not\preceq_s \perp_{B,L}(e^-)$. Then $H \in \mathcal{H}_{L,C \cup C'}$. $C'$ is sound regardless of whether $H$ is a solution to $P$.

Suppose $H \preceq_s \perp_{B,L}(e^-)$. Then $H \notin \mathcal{H}_{L,C \cup C'}$. To show $C'$ is sound, it is sufficient to show that $H$ is not a solution to $P$. Since $\preceq_s$ is sound, $H$ must be a generalization of $\perp_{B,L}(e^-)$. In Proposition 2, it was shown that $H$ cannot a solution to $P$ and thus $C'$ is sound. □

Proposition 4 means that negative bottom preprocessing is sound as long as a sound generalization relation is used. This means the generalization relation is free to be incomplete and can misclassify some generalizations as non-generalizations. The most extreme version of this might be a generalization relation that returns false, i.e. it assumes nothing is a generalization.

### 3.3.2 Positive Bottom Preprocessing

Proposition 3 can be extended to show that positive bottom preprocessing is still subsumption-sound if it uses a complete generalization relation $\preceq_c$.

**Proposition 5.** *Suppose $P = (B, L, C, E^+, E^-)$ is an LFF problem input. Suppose $\preceq_c$ is a subsumption-complete generalization relation. Suppose $C'$ is a a set of hypothesis constraints generated by positive bottom preprocessing with $\preceq_s$ defining generalization. Then $C'$ is subsumption-sound.*

*Proof.* Suppose $P$, $\preceq_c$, and $C'$ are as stated. Suppose $e^+ \in E^+$ and $H \in \mathcal{H}_{L,C}$.

Suppose $H \preceq_c \perp_{B,L}(e^+)$. Then $H \in \mathcal{H}_{L,C \cup C'}$. $C'$ is subsumption sound regardless of whether $H$ is a solution to $P$.

Suppose $H \not\preceq_c \perp_{B,L}(e^+)$. Then $H \notin \mathcal{H}_{L,C \cup C'}$. Since $\preceq_c$ is subsumption-complete, $H$ is not a generalization of $\perp_{B,L}(e^+)$. In Proposition 3 it was shown that if $H$ is a subsumption complete solution to $P$ then $H$ must a generalization of $\perp_{B,L}(e^+)$. Thus $H$ is not a subsumption complete solution to $P$ and $C'$ is sound. □

From Proposition 5 it suffices to use a complete definition of generalization to ensure positive bottom preprocessing does not prune subsumption-complete solutions. In the extreme case a complete generalization relation might return true, i.e. it assumes everything is a generalization. Chapter 4 shows that useful complete generalization relations are harder to compute than sound but incomplete generalization relations. Experiments in chapter 5 demonstrate that even hypothesis constraints generated by bottom preprocessing of positive examples using an incomplete definition of generalization still prune very few solutions on a variety of problems.

# Chapter 4

# ⊥-Popper Implementation

⊥-Popper implements bottom preprocessing for the meta-level ILP system Popper [7]. This chapter describes the implementation of ⊥-Popper. While ⊥-Popper is a specific implementation, the techniques described here could be adapted to other meta-level ILP systems that support hypothesis constraints.

**Remark.** *The background knowledge in this chapter is assumed to be written in Prolog. It is not guaranteed to have a finite Herbrand base.*

Prolog also supports negation as failure. This can cause Prolog programs to be non-monotonic. It is assumed that the BK is written such that it is a definite program.

## 4.1 Overview

⊥-Popper takes an LFF problem input of the form $(B, L, C, E^+, E^-)$ as given in Definition 11. It generates a set of additional hypothesis constraints $C'$ using bottom preprocessing. Then it passes a new LFF problem input $(B, L, C \cup C', E^+, E^-)$ to Popper for solving and returns the result. Thus ⊥-Popper functions as a wrapper around Popper.

The interface between ⊥-Popper and Popper is defined by the hypothesis language $\mathcal{L}$ introduced in Definition 9. All together this means that ⊥-Popper is agnostic to the implementation details of Popper. A brief explanation of how Popper works is given in Chapter 6.

### 4.1.1 High Level Algorithm

A simplified version of the ⊥-Popper algorithm is shown in Algorithm 1. ⊥-Popper consists of four distinct components which are described in separate sections:

**Algorithm 1** The ⊥-Popper algorithm.
___
1: $g \leftarrow$ GEN_RELS
2: $h \leftarrow$ HYP_CONSTRAINTS
3: **function** ⊥-POPPER($e^+, e^-, B, L, C, P$)
4:     $b \leftarrow$ BC_PROG($e^+, e^-, B, L, P$)
5:     $b' \leftarrow$ BC_ENC($b, P$)
6:     $C' \leftarrow C \cup b' \cup g \cup h$
7:     **return** POPPER($e^+, e^-, B, L, C', P$)
8: **end function**
___

1. `hyp_constraints`: Positive and negative bottom preprocessing hypothesis constraints in $\mathcal{L}$ which depend on a generalization relation.

2. `gen_rels`: A generalization relation in $\mathcal{L}$ written in terms of bottom clause encodings.

3. `bc_enc`: A program which generates encodings in $\mathcal{L}$ of a set of bottom clauses.

4. `bc_prog`: A Prolog program which computes the bias consistent bottom clauses of all examples in an ILP problem.

In addition to LFF problem input, ⊥-Popper expects parameters $P$ of the form (`max_vars`, `max_literals`, `max_clauses`). These parameters limit respectively the number of variables in a hypothesis, the number of literals in a clause, and the number of clauses in a hypothesis. Only `max_vars` is used by ⊥-Popper. The remaining arguments are passed on to Popper without inspection.

### 4.1.2   $\mathcal{L}$: Hypothesis Language

Hypothesis constraints in Popper are written in ASP. This gives ⊥-Popper the full expressivity of ASP in which to write hypothesis constraints, generalization relations, and bottom clause encodings.

Popper encodes guessed hypotheses in ASP using the following predicate symbols: `head_literal/2`, `body_literal/2`. ⊥-Popper prunes the hypothesis space by encoding constraints on facts involving these predicate symbols and `clause/1`. This section briefly describes the encoding. The details of the full encoding can be found in chapter 4 of the Popper paper [7].

Suppose $H$ is the current hypothesis guessed by Popper and $H$ consists of clauses $\{C_1, C_2, \cdots, C_n\}$. Then $H$ corresponds to a set of facts $F$ in $\mathcal{L}$ encoding each $C_i$. It must be the case that $n$ is less than `max_clauses` and `clause`($i$) $\in F$ for each $C_i$ automatically. Suppose $X = [V_0, V_1, \cdots, V_n]$ is a list of the variables in $C_i$ sorted by their variable name. Then $\theta = \{V_0/0, V_1/1, \cdots, V_n/n\}$ is a substitution of each $V_i$ for its index $i$ in $X$. Suppose $C_i\theta$ is of the form $h \leftarrow b_1, b_2, \cdots, b_n$. Then $C_i\theta$ is encoded in $F$ as follows:

- If $h$ is of the form $p(x_0, x_1, \cdots, x_n)$ then $\texttt{head\_literal}(\text{id}_{C_i}, p, n, (x_0, x_1, \cdots, x_n)) \in F$.

- For each $b_i$ of the form $p(x_0, x_1, \cdots, x_n)$, $\texttt{body\_literal}(\text{id}_{C_i}, p, n, (x_0, x_1, \cdots, x_n)) \in F$.

**Example 4.1.** Example 2.5 already demonstrated this encoding. It is briefly repeated here. Suppose the hypothesis $H$ is:

```
palindrome(A) :- middle(A, B), empty(B).
```

Then the encoding in $\mathcal{L}$ of $H$ is:

```
clause(0).
head_literal(0, palindrome, 1, (0,)).
body_literal(0, middle, 2, (0, 1)).
body_literal(0, empty, 1, (1,)).
```

### 4.1.3 Bottom Clause Encoding

$\bot$-Popper encodes multiple *bottom clause variants* for each bottom clause. The details of how bottom clause variants are chosen is given in Section 4.4. Here bottom clause variants are defined and an explanation of how they are encoded is given.

**Definition 34** (Bottom clause variant)**.** Suppose $P = (B, L, C, E^+, E^-)$ is an LFF problem input. Suppose $e \in E^+ \cup E^-$. A *variant* of the bottom clause $\bot_{B,L}(e)$ is a bias consistent clause $D$ such that $D$ subsumes $\bot_{B,L}(e)$.

A bottom clause variant can be encoded in $\mathcal{L}$ as a set of facts in ASP. Suppose $P = (B, L, C, E^+, E^-)$ is an LFF problem input and $D$ is a variant of $\bot_{B,L}(e)$ for some example $e \in E^+ \cup E^-$. Suppose $id_{\bot(e)}$ is a unique identifier for $\bot_{B,L}(e)$. Suppose $id_D$ is a unique identifier for $D$. Suppose $X = [V_0, V_1, \cdots, V_n]$ is a list of the variables in $D$ sorted by their variable name. Then $\theta = \{V_0/0, V_1/1, \cdots, V_n/n\}$ is a substitution of each $V_i$ for its index $i$ in $X$. Suppose $D\theta$ is of the form $h \leftarrow b_1, b_2, \cdots b_n$. Then $D\theta$ can be encoded as a set of ground facts $F$ in the hypothesis language $\mathcal{L}$ as follows:

- If $e \in E^+$, $\texttt{pos\_bottom\_clause}(id_{\bot(e)}) \in F$. Otherwise when $e \in E^-$, $\texttt{neg\_bottom\_clause}(id_{\bot(e)}) \in F$.

- `bottom_clause_variant`$(id_{\perp(e)}, id_D) \in F$.

- If the atom $h$, the head of $D\theta$, is of the form $p(x_0, x_1, \cdots, x_n)$ then `bottom_clause_head`$(id_D, p, (x_0, x_1, \cdots, x_n)) \in F$. Since $D$ is definite, $h$ always exists. Note that $p$ is a predicate symbol and $(x_1, x_2, \cdots, x_n)$ is a tuple of integers.

- For each atom $b_i$ of the form $p(x_0, x_1, \cdots, x_n)$ in the body of $D\theta$, `bottom_clause_body`$(id_D, p, (x_0, x_1, \cdots, x_n)) \in F$.

**Example 4.2.** Suppose $\perp_{B,L}(e)$ is the bottom clause of the negative example `palindrome([a, t])` from Example 1.2:

```
palindrome(A) :- first(A,B), last(A,C), middle(A,D), empty(D),
   length(A,F), length(D,G), one(E).
```

Suppose the variant $D$ is equal to the entire bottom clause $\perp_{B,L}(e)$. The unique identifier $id_{\perp(e)}$ might be `neg_1` and $id_D$ might be `neg_1_1`. Then $D\theta$ is:

```
palindrome(0) :- first(0,1), last(0,2), middle(0,3), empty(3),
   length(0,5), length(3,6), one(4).
```

The encoding of $D$ in $\mathcal{L}$ is then:

```
neg_bottom_clause(neg_1).
bottom_clause_variant(neg_1, neg_1_1).
bottom_clause_head(neg_1_1, palindrome, (0,)).
bottom_clause_body(neg_1_1, first, (0,1,)).
bottom_clause_body(neg_1_1, last, (0,2,)).
bottom_clause_body(neg_1_1, middle, (0,3,)).
bottom_clause_body(neg_1_1, empty, (3,)).
bottom_clause_body(neg_1_1, length, (0,5,)).
bottom_clause_body(neg_1_1, length, (3,6,)).
bottom_clause_body(neg_1_1, one, (4,)).
```

## 4.2 `hyp_constraints`: Hypothesis Constraints

The hypothesis constraints for $\perp$-Popper are straightforward encodings into ASP of bottom preprocessing as given in Definition 31. This ASP code is given in Algorithm 2. The symbol _ is an anonymous variable. If `generalizes_bottom_clause(_, D)` holds then there is some clause `Cl` in the current guessed hypothesis such that `generalizes_bottom_clause(Cl, D)`. In other words if `generalizes_bottom_clause(_, D)` holds, the current hypothesis $H$ is a generalization of $D$.

**Algorithm 2** ASP code for the component `hyp_constraints`

```
1  % The following is constraint C⁻
2  :-
3      neg_bottom_clause(D),
4      generalizes_bottom_clause(_, D).
5
6  % The following is constraint C⁺
7  :-
8      pos_bottom_clause(D),
9      not generalizes_bottom_clause(_, D).
```

**Proposition 6.** *Suppose $P = (B, L, C, E^+, E^-)$ is an LFF problem input. Suppose $H \in \mathcal{H}_{L,C}$ is the current hypothesis encoded by $\mathcal{L}$. Suppose `neg_bottom_clause(D)` holds if and only if $D$ is $\bot_{B,L}(e^-)$ for some $e^- \in E^-$. Suppose `pos_bottom_clause(D)` holds if and only if $D$ is $\bot_{B,L}(e^+)$ for some $e^+ \in E^+$. Then the following are true:*

- *If `generalizes_bottom_clause/2` is subsumption-sound, hypothesis constraint $C^-$ is sound.*

- *If `generalizes_bottom_clause/2` is subsumption-complete, hypothesis constraint $C^+$ is subsumption-sound.*

- *If `generalizes_bottom_clause/2` is both subsumption-sound and subsumption-complete, the hypothesis constraints in `hyp_constraints` are subsumption-sound.*

*Proof.* Suppose $P$, `neg_bottom_clause/1`, `pos_bottom_clause/1`, and $H$ are as stated.

Suppose `generalizes_bottom_clause/2` is subsumption-sound. Suppose $D$ is $\bot_{B,L}(e^-)$ for some $e^- \in E^-$. Then `neg_bottom_clause(D)` holds. Suppose `generalizes_bottom_clause(_, D)` holds. This means $C^-$ is violated. So $H \notin \mathcal{H}_{L,C\cup C^-}$. Inversely, since $H \in \mathcal{H}_{L,C}$, if $H \notin \mathcal{H}_{L,C\cup C^-}$ then $C^-$ must be violated. So `generalizes_bottom_clause(_, D)` must hold. Taking the contrapositive, we have that for all $e^- \in E^-$, $H \in \mathcal{H}_{L,C\cup C^-}$ if and only if $\neg$ `generalizes_bottom_clause(_, D)`. So by Proposition 4, $C^-$ is sound.

Suppose `generalizes_bottom_clause/2` is subsumption-complete. Very similar reasoning shows that $C^+$ is a constraint such that for all $e^+ \in E^+$, $H \in \mathcal{H}_{L,C\cup C^+}$ if and only if `generalizes_bottom_clause(_, D)`. So, by Proposition 5, $C^+$ is subsumption-sound.

Suppose `generalizes_bottom_clause/2` is both subsumption-sound and subsumption-complete. Then $C^-$ is sound and $C^+$ is subsumption-sound. Suppose $H$ is a subsumption complete solution to $P$. Since $C^-$ is sound and $H$ is a solution to $P$, $H \in \mathcal{H}_{L,C \cup C^-}$. Since $C^+$ is subsumption-sound and $H$ is a subsumption-complete solution to $P$, $H \in \mathcal{H}_{L,C \cup C^+}$. So $H \in \mathcal{H}_{L,C \cup C^- \cup C^+}$ and the hypothesis constraints in `hyp_constraints` are subsumption-sound. $\square$

Proposition 6 provides a specification for the rest of the implementation of $\bot$-Popper. Ideally the following things should be true:

1. `neg_bottom_clause(D)` holds whenever $D$ is the bottom clause of a negative example.

2. `pos_bottom_clause(D)` holds whenever $D$ is the bottom clause of a positive example.

3. `generalizes_bottom_clause(_, D)` is a sound and complete generalization relation. It suffices that `generalizes_bottom_clause(C, D)` is a subsumption-sound and subsumption-complete generalization relation between clauses.

Item 3 is computationally challenging since subsumption checking is NP-complete. The next section on `gen_rels` describes a subsumption-sound implementation of `generalizes_bottom_clause/2`. The section on `bc_enc` shows how to increase the coverage of `generalizes_bottom_clause/2` toward subsumption-completeness by encoding many variants of the bottom clause.

## 4.3 `gen_rels`: Generalization Relations

The component `gen_rels` encodes a generalization relation between clauses in ASP. If a clause $C_1$ generalizes a clause $C_2$ then there is a substitution $\theta$ such that $C_1\theta \subseteq C_2$. Suppose that substitution is ignored and it is assumed that $\theta = \{\}$. Then $C_1\theta = C_1$. If $C_1 \subseteq C_2$, it must be the case that for all literals $l \in C_1$, $l \in C_2$. This test turns out to be inefficient to encode in ASP. Instead, it is easier to test whether there exists some literal $l$ in $C_1$ such that $l \notin C_2$. For this reason `gen_rels` tests whether $C_1 \subseteq C_2$ by checking if $\neg(C_1 \nsubseteq C_2)$.

A generalization relation which ignores substitution is subsumption-sound. Suppose $C_1$ and $C_2$ are clauses and `sound_gen/2` is a generalization relation such that

**Algorithm 3** ASP code for the component `gen_rels`

```
1  generalizes_bottom_clause(Cl, D) :-
2      clause(Cl),
3      bottom_clause_variant(D, V),
4      not does_not_generalize_bottom_clause_variant(Cl, V)
5
6  does_not_generalize_bottom_clause_variant(Cl, V):-
7      bottom_clause_variant(_, V),
8      head_literal(Cl, P, _, Vars),
9      not bottom_clause_head(V, P, Vars).
10 does_not_generalize_bottom_clause_variant(Cl, V):-
11     bottom_clause_variant(_, V),
12     body_literal(Cl, P, _, Vars),
13     not bottom_clause_body(V, P, Vars).
```

`sound_gen`$(C_1, C_2)$ if and only if $\neg(C_1 \not\subseteq C_2)$. Then whenever `sound_gen`$(C_1, C_2)$ holds, $C_1\theta \subseteq C_2$ for the substitution $\theta = \{\}$. Thus $C_1$ generalizes $C_2$.

The generalization relation `sound_gen/2` is far from complete though. The next section describes a component `bc_enc` which generates a set of bottom clause variants from a bottom clause. Suppose `bc_enc` is a function such that if $\bot$ is a bottom clause, $C$ is a clause and $\theta$ is some substitution, $C\theta \subseteq \bot$ if and only if $C \subseteq \bot_1$ for some bottom clause variant $\bot_1 \in $ `bc_enc`$(\bot)$. Say that `bc_enc` is complete in this case. Suppose `complete_gen/2` is a generalization relation such that `complete_gen`$(C, \bot)$ if and only if $\neg(C \not\subseteq \bot_1)$ for some $\bot_1 \in $ `bc_enc`$(\bot)$. Then `complete_gen/2` is a subsumption-complete generalization relation if and only if `bc_enc` is a complete encoding of bottom clause variants.

The ASP code for the component `gen_rels` is given in Algorithm 3. When a bottom clause is encoded as a single variant equal to itself, `generalizes_bottom_clause/2` is equivalent to `sound_gen/2`. Thus `generalizes_bottom_clause/2` is subsumption-sound in that case. When `bc_enc` is complete then `generalizes_bottom_clause/2` is equivalent to `complete_gen/2`. Thus the subsumption-completeness and subsumption-soundness of `gen_rels` depend on the bottom clause variants encoded by `bc_enc`.

## 4.4 `bc_enc`: Bottom Clause Encoding

`bc_enc` is a component which generates multiple bottom clause variants from each bottom clause and encodes them in $\mathcal{L}$. Ideally this approach can ensure that `gen_rels` is both subsumption-sound and subsumption-complete.

When $\perp_B(e)$ for a fact $e$ is lifted to variables, it is assumed that any co-occurences of the same ground term in $\perp_B(e)$ should be lifted to the same variable. For instance, a ground atom `middle([], [])` might be lifted to the atom `middle(A, A)`. If $\perp_{B,L}(e)$ is then `palindrome(A):- middle(A, A)`, the clause `palindrome(A):- middle(A, B)` generalizes $\perp_{B,L}(e)$ with the substitution $\theta = \{B/A\}$.

The original Progol implementation handled this case by performing a technique called *variable splitting* during its search [20]. Later systems such as Aleph and ATOM perform variable splitting on the bottom clause instead [1].

To describe variable splitting it is first necessary to define the *direction* of a variable in a literal. Suppose $L$ is a language bias. Suppose $C$ is a clause. Suppose $V_i$ is a variable occurring in a literal $a \in C$ with form $p(V_1, V_2, \cdots, V_n)$. Suppose there is a direction declaration of the form `direction`$(p, (D_1, D_2, \cdots, D_n)) \in L_d$. Then $V_i$ has *direction* $D_i$ in $a$. As per the definition of direction declarations, $D_i$ is either `in` or `out`.

*Variable splitting* a clause $C$ under $L$ generates a clause $C_{VS}$ according to the following algorithm:

1. Initialize $C_{VS}$ to be $C$.

2. Suppose $C_{VS}$ has the form $h \leftarrow b_1, b_2, \cdots, b_n$. A variable $X$ is splittable if $X$ occurs in $h$ or $X$ occurs with direction `out` in some $b_i$. Find two co-occurrences of a splittable variable $V$ in two literals $a_1$ and $a_2$ in $C_{VS}$. If $a_1 = a_2$ then $V$ must occur as arguments $i$ and $j$ where $i \neq j$. If no co-occurences of splittable variables exist, return $C_{VS}$.

3. Replace the occurrence of $V$ in $a_2$ with a new variable $V'$ to create a literal $a_2'$. Replace $b$ in $C_{VS}$ with $b'$. Also add $V'$ to $[\![V]\!]$, the equivalence class of $V$.

4. For all literals $b_i$ with $V$ having direction `in` in $b_i$, replace $V$ with $V'$ in a copy $b_i$ to create $b_{n+1}$. Add $b_{n+1}$ to $C_{VS}$. Return to step 2.

Variable splitting ensures that all variables with direction `out` in $C_{VS}$ are unique. It can also add a large number of extra literals to the bottom clause when a copied literal is split further.

Aleph and ATOM add equality literals $V = V'$ to the bottom clause if variable $V'$ was added to the equivalence class of $V$ during variable splitting. Thus the example $C =$`palindrome(A):- middle(A, A)` might become $C_{VS} =$`palindrome(A):- middle(A, B), A = B` after variable splitting. Then any subset of the literals in $C_{VS}$ is logically equivalent to a generalization of $C$.

`bc_enc` takes a slightly different approach. Suppose $C$ is a clause and $C_{VS}$ is $C$ after variable splitting. Suppose $V = \{V_0, V_1, \cdots, V_n\}$ is the set of all unique variables in $C_{VS}$. `bc_enc` encodes a bottom clause variant for each possible substitution $\theta$ of the variables in $V$ to a subset of the variables in $V$ where $V_i/V_k$ and $V_j/V_k$ in $\theta$ only if $V_j \in [\![V_i]\!]$. In other words $V_i$ and $V_j$ can only map to the same variable in $\theta$ if they were added to the same equivalence class during variable splitting.

When working on this, I was under the impression that variable splitting would make `bc_enc` complete for subsumption. Specifically, that the following would be true. Suppose $C$ and $D$ are bias consistent clauses for some language bias $L$. Suppose $D_{VS}$ is $D$ after undergoing variable splitting. If $C\theta \subseteq D$ for some substitution $\theta$ then there exists a substitution $\theta'$ such that $C \subseteq D_{VS}\theta'$.

However, there is a counterexample to this in a paper by Tamaddoni-Nezhad and Muggleton discussing the completeness of Progol's refinement operator [28]. Suppose $D$ is `palindrome(A):- middle(A, A)` and $D_{VS}$ is then `palindrome(A):- middle(A, B)` according to the variable splitting algorithm. Suppose $D' =$ `palindrome(A):- middle(A, B), middle(B, A)`. Then $D'$ subsumes $D$ but $D' \not\subseteq D_{VS}\theta$ for any substitution $\theta$. It can be argued that $D'$ is redundant because it is logically equivalent to $D$ and thus it would never appear in an optimal solution.

This example still brings into question the completeness of this bottom clause encoding approach. I was unable to complete a proof I felt was strong enough to include here. Experiments in Chapter 5 show that if `generalizes_bottom_clause/2` is not a complete generalization relation with this implementation of `bc_enc`, it at least works well in practice. In the limitations section of Chapter 7, I briefly describe several ways the implementation of a generalization relation for positive bottom preprocessing could be changed to overcome this possible limitation.

`bc_enc` encodes bottom clause variants by the following steps:

1. For each bottom clause $\bot_{B,L}(e)$ where $e \in E^+ \cup E^-$, generate the variable split version of the bottom clause, $\bot_{B,L}^{VS}(e)$.

2. Find the set $T$ of all possible substitutions of $\bot_{B,L}^{VS}(e)$.

3. Encode a bottom clause variant in $\mathcal{L}$ by computing $\bot_{B,L}^{VS}(e)\theta$ for each substitution $\theta \in T$.

In many cases, with enough examples, a target hypothesis can be found without variable splitting. However, there are certain problems for which a target hypothesis cannot be found without variable splitting [28]. $\bot$-Popper can be run with or without variable splitting enabled. If variable splitting is disabled, $\perp_{B,L}^{VS}(e) = \perp_{B,L}(e)$ and the substitutions in $T$ are one-to-one. Experiments in Chapter 5 show that variable splitting can dramatically increase learning times and is frequently unnecessary to find an optimal solution.

### 4.4.1 Optimization of Substitutions

Suppose a variable split bottom clause $\perp_{VS}$ contains $n$ variables $(V_0, V_1, \cdots, V_n)$. Then there are $\binom{2n}{n}$ possible substitutions of the $n$ variables to some subset of $(V_0, V_1, \cdots, V_n)$. There are several optimizations that can reduce the number of substitutions that need to be considered.

The restriction described in the previous section must be obeyed for soundness. If $V_i/V_k$ and $V_j/V_k$ are in some substitution $\theta$ then $V_i$ and $V_j$ must be assigned to the same equivalence class during variable splitting. In the worst case, all variables are in the same equivalence class and this does not change the number of variants generated.

Suppose `max_vars` is $j$ for some problem. Suppose `body_literal(Cl, P, N, Vars)` is a body literal guessed as part of a hypothesis $H$ by Popper. Then `Vars` is a tuple of integers. Call these *variable indices*. Popper's guessing is bounded such that it is not possible that there exists a variable index $i \in$ `Vars` such that $i \geq j$. Then it is not necessary to consider any substitution $\theta$ where the encoding of $\perp\theta$ would contain a variable index greater than $j - 1$. This reduces the number of substitutions that need to be considered to $\binom{n+j}{j}$. In fact, as a further optimization, all literals $l \in \perp_{VS}$ containing a variable $V_i$ such that $V_i \notin \theta$ can be dropped from $\perp_{VS}\theta$.

Suppose `head_literal(C, f, N, Vars)` is the head of the current hypothesis $H$ in Popper. The variables in `Vars` are always assigned fixed indices $0, 1, \cdots, N$. Thus, the variables in the head of a bottom clause variant $\perp$ can also be fixed. Substitutions which permute these variables do not need to be considered. This reduces the number of substitutions that need to be considered for a bottom clause with $n$ variables, `max_vars` $j$, and a head predicate symbol with arity $h$ to $\binom{n+j-h}{j-h}$. Since $j$ is frequently defaulted to 5 in Popper and $h$ is at least 1 and very frequently 2 or greater, the number of variants can become manageable with these optimizations.

Finally, taking a subset of the bottom clause might cause a bottom clause variant to no longer be bias consistent. Specifically, it can violate direction declarations. Thus, any substitution which would generate a variant that is not bias consistent can be ignored. The reduction this causes in the number of substitutions considered is difficult to compute since it depends heavily on the language bias provided.

All of these optimizations are implemented in ⊥-Popper by an ASP program. Given a clause and a set of constraints encoding these restrictions, the ASP program generates all necessary substitutions of size `max_vars`. The code that initializes and runs this ASP program is written in Python. It can be found in Appendix C

Suppose some bottom clause variant $D$ contains $k$ literals. Then the encoding of $D$ in $\mathcal{L}$ contains $k+1$ facts. There is also one fact encoded per $e \in E^+ \cup E^-$. Suppose $V$ is the set of bottom clause variants generated by `bc_enc` and $m$ is the maximum number of literals in any bottom clause variant in $V$. Then the maximum number of facts encoded in $\mathcal{L}$ by bottom clause encoding is $(m+1)|V| + |E^+ \cup E^-|$.

On a small technical note, a practical optimization in the implementation of Popper ensured that Popper never generated hypotheses in which the same variable occurred twice in a literal. This actually compromises the optimality of Popper. For instance, with this optimization, Popper was unable to find the simplest program for a palindrome: `palindrome(A):- reverse(A, A)`. This optimization would have made variable splitting unnecessary in many cases and possibly affected the experiments in the next chapter. So this optimization was disabled in ⊥-Popper and the version of Popper used for experiments in Chapter 5.

## 4.5 `bc_prog`: Bottom Clause Generation

So far this chapter has shown how subsumption-sound hypothesis constraints can be generated from the bottom clauses of examples. It remains to describe the last component, `bc_prog`, which constructs a bottom clause for each example in an LFF problem input.

When the BK of the problem is a Datalog program, the bottom clause of an example can be computed by finding the least Herbrand model of $B \cup \neg e$ [8]. Since $B \cup \perp_B(e) \models e$ it must be the case that $B \cup \neg e \models \neg \perp_B(e)$. If the least Herbrand model of $B \cup \neg e$ is a set of ground facts $b_1 \wedge b_2 \wedge \cdots \wedge b_n$ then $\perp_B(e)$ is $\neg(\neg e \wedge b_1 \wedge b_2 \wedge \cdots \wedge b_n)$. This means $\perp_B(e)$ is the clause $e \leftarrow b_1, b_2, ...b_n$. Note that the use of negation in this manner is only valid because $e$ and all $b_i$ are facts. This algorithm could be called bottom clause construction with *forward chaining*, or $\text{BC}_{FC}$ in brief, since the least Herbrand model of a Datalog program can be computed using a forward chaining algorithm [8].

Two problems exist with $\text{BC}_{FC}$ when the BK is written in Prolog. Since Prolog is not guaranteed to have a finite Herbrand base, the computation of the least Herbrand model can be infinite. Forward chaining can also fail if clauses in the BK are not *range-restricted*, i.e they contain variables in the head that do not exist in the body. Since Popper allows Prolog BK, ⊥-Popper instead uses the bottom clause construction algorithm given in the original inverse entailment paper [20], call it $BC_{IE}$. $BC_{IE}$ has the benefits of being more efficient than $BC_{FC}$ and computing the bias consistent bottom clause of an example directly. However, $BC_{IE}$ is significantly more complex than $BC_{FC}$.

`bc_prog` implements $BC_{IE}$ in approximately 200 lines of Prolog. Since $BC_{IE}$ is well studied and the proof of its correctness is given in the original paper, a complete description is omitted here. The code used to perform bottom clause construction in ⊥-Popper can be found in Appendix B. Appendix A also provides a program which converts the LFF language bias given in Definition 6 into mode declarations. The fact that `bc_prog` constructs bias consistent bottom clauses then follows directly from the proof that $BC_{IE}$ constructs mode consistent bottom clauses.

$BC_{IE}$ ensures its construction of a bottom clause will terminate by bounding it using a concept called *variable depth*.

**Definition 35** (Variable Depth). If $C$ is a clause and $V = \{v_1, v_2, \cdots, v_n\}$ is the set of unique variables occurring in C, then the *variable depth* of a variable $v_i \in V$, $\text{depth}(v_i)$, is defined as follows:

- $\text{depth}(v_i)$ is 0 if $v_i$ occurs in the head of $C$.

- Otherwise, let $A_{v_i}$ be the set of atoms in $C$ in which $v_i$ occurs. Let $U = \{u_1, u_2, \cdots, u_n\}$ be the set of variables occurring in the atoms in $A_{v_i}$. Suppose $d$ is the minimum, for all $u_i \in U$ such that $u_i \neq v_i$, of $\text{depth}(u_i)$. Then $\text{depth}(v_i)$ is $d + 1$.

The computational complexity of bottom clause construction is exponential in the value of `max_variable_depth` [20]. However, setting the value too low can exclude literals from the bottom clause. Aleph [27], a popular implementation of Progol in Prolog, defaults `max_variable_depth` to 3.

⊥-Popper defaults `max_variable_depth` for $BC_{IE}$ to `max_vars` $-1$ in `bc_prog`. The following proposition shows why this is a sound choice.

**Proposition 7.** *If $C$ is a clause and $V = \{v_1, v_2, \cdots, v_n\}$ is the set of unique variables occurring in C, the maximum variable depth of any variable $v_i \in V$ is $|V| - 1$.*

*Proof.* Suppose $C$ and $V$ are as stated and $|V| = k$. Suppose, for contradiction, $v_i \in V$ has depth$(v_i) = k$. Then there must be some chain of unique variables $L = l_0, l_1, \cdots, l_{k-1}$ such that for each $j$ between 0 and $k - 1$, depth$(l_j) = j$, $l_j \in V$ and $l_j \neq v_i$. The length of $L$ is $k$ which means $C$ must contain $k+1$ unique variables. This is a contradiction. $\square$

Suppose $M$ is the set of mode declarations in a language bias $L$, $j^+$ and $j^-$ are the maximum number of input and output variables respectively in any mode declaration in $M$, $r$ is the maximum recall in any mode declaration in $M$, and $i$ is `max_variable_depth`. As stated in the original paper [20], the computational complexity of BC$_{IE}$ is bounded by $(j^+j^-r|M|)^{ij^+}$. Thus bottom clause construction in ⊥-Popper can theoretically scale poorly in the size of the language bias or the value of `max_vars`. In practice, this is rarely a problem as will be shown in chapter 5. Usually `max_vars` defaults to 5 in Popper problems so $i = 4$. Typically, $j^+ \leq 2$. Since recall defaults to $*$, $r$ is potentially infinite but the term $j^+j^-r$ is also bounded by the number of true atoms in the least Herbrand model at depth $i$ which is typically fairly small.

## 4.6 Recursion Constraints

Proposition 7 proves that no hypothesis Popper produces can contain a clause with a variable having variable depth greater than `max_vars - 1`. However, if a hypothesis can self-resolve, then the result of multiple self-resolutions might produce a clause with variables having higher variable depths than the total number of variables in the theory. For this reason, ⊥-Popper prunes recursive theories more cautiously than other theories.

Algorithm 4 contains additional hypothesis constraints which loosen `generalizes_bottom_clause/2` in the case of problems which allow recursion. Suppose $C$ is a clause of the form $h \leftarrow b_1, \cdots, b_n$ in which some $b_r$, with $1 \leq r \leq n$, has the same predicate symbol as $h$. Then $C$ is a recursive clause. Suppose $k$ is the minimum variable depth of the input variables in $b_r$. Say that the literal $b_i$ *precedes recursion* if depth$(V_j) < k$ for all output variables $V_j$ in $b_i$.

The constraints in Algorithm 4 add an additional case to `generalizes_bottom_clause/2` when the bottom clause is constructed from a positive example and the problem supports recursion. In this case, for a clause $C$ and a positive bottom clause variant $D$, `generalizes_bottom_clause(C, D)` holds if the subset of $C$ consisting of literals that precede recursion, call it $C_r$, is a subset of $D$. This makes

**Algorithm 4** ASP code which loosens hypothesis constraints in the presence of recursion.

```
1  generalizes_bottom_clause(Cl, E) :-
2      recursive,
3      clause(Cl),
4      pos_bottom_clause(E),
5      bottom_clause_variant(E, V),
6      not does_not_recursively_generalize_bottom_clause_variant(Cl, V).
7
8  does_not_recursively_generalize_bottom_clause_variant(Cl, V):-
9      bottom_clause_variant(_, V),
10     precedes_recursion(Cl, P, Vars),
11     not bottom_clause_body(V, P, Vars).
12
13 precedes_recursion(Cl, P, Vars):-
14     head_literal(Cl, PI, AI, _),
15     body_literal(Cl, PI, AI, VarsI),
16     in_var_is_out_var(Cl, PI, VarsI, P, Vars).
17 precedes_recursion(Cl, P, Vars):-
18     precedes_recursion(Cl, PI, VarsI),
19     in_var_is_out_var(Cl, PI, VarsI, P, Vars).
20
21 in_var_is_out_var(Cl, PI, VarsI, P, Vars) :-
22     var_direction(PI, VarsI, in, Var),
23     body_literal(Cl, P, _, Vars),
24     var_direction(P, Vars, out, Var).
```

generalizes_bottom_clause/2 an unsound generalization relation during positive bottom preprocessing of recursive problem. However, generalizes_bottom_clause/2 can still be a complete generalization relation in this case. Proposition 5 showed this this condition is sufficient for positive bottom preprocessing to remain subsumption-sound.

Note that Algorithm 4 requires some definition of the undefined predicate symbol var_direction/4. Suppose var_direction(P, Vars, Dir, Var) holds if Vars has the form $(V_1, \cdots, V_n)$, Var is some $V_i$ and the direction declaration of the form direction(P, $(D_1, \cdots, D_n)$) in the language bias of the problem has $D_i = $ Dir. Then Algorithm 4 encodes this restriction.

# Chapter 5

# Experimental Results

Three major claims have been made about how bottom preprocessing performs in practice. The experiments in this chapter test these claims by answering the following questions:

- **Q1**: Can bottom preprocessing significantly reduce the number of programs a meta-level ILP system explores?

- **Q2**: Can bottom preprocessing reduce learning times of a meta-level ILP system on hard problems?

- **Q3**: Can bottom preprocessing improve the scaling of a meta-level ILP system as the amount of irrelevant BK increases?

Beyond these primary questions, there are several more specific claims about the implementation of $\bot$-Popper which are worth evaluating experimentally:

- **Q4**: Are positive and negative bottom preprocessing similarly effective at reducing the hypothesis space of LFF problems?

- **Q5**: Does unsound bottom preprocessing frequently preserve optimal solutions in practice?

- **Q6**: Does variable splitting significantly increase learning times?

To evaluate questions **Q4** through **Q6**, we use several different versions of $\bot$-Popper:

- $\bot$-Popper is the default version. It uses both positive and negative example bottom preprocessing. However, it performs no variable splitting. This version may prune subsumption-complete solutions.

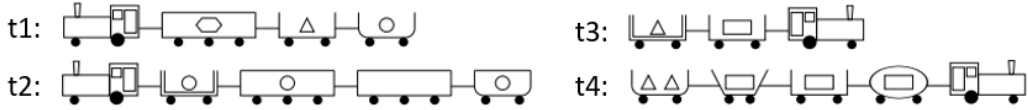

Figure 5.1: Examples from the Michalski trains problem.

- ⊥-Popper$_{VS}$ is ⊥-Popper with variable splitting enabled. This version may prune solutions but does not prune subsumption-complete solutions.

- ⊥-Popper$_-$ performs only negative example bottom preprocessing. This version does not prune solutions.

- ⊥-Popper$_+$ performs only positive example bottom preprocessing. This version may prune subsumption-complete solutions.

All of the experiments in this chapter were run on an Amazon EC2 c5.4xlarge instance with 16 vCPUs and 32 GB of memory.

## 5.1 Trains Problems

Michalski's train problem [13] is a classic problem in machine learning. Figure 5.1 shows four example trains in the problem. Trains `t1` and `t2` are westbound, facing left, while `t3` and `t4` are eastbound, facing right. The goal for a trains problem is to find a relation `f(Train)` which holds for eastbound trains but not for westbound trains. Finding such a relation was the basis of a machine learning competition held in 1994. These images were taken from that competition [18].

Since the instance of the trains problem in Figure 5.1 is simple, the reader might be able to find a solution with a quick visual inspection. All eastbound trains have a car carrying a rectangular load while none of the westbound cars carry a rectangular load. Popper, given an encoding of this problem as an LFF problem quickly finds the following solution:

`f(A) :- has_car(A, B), has_load(B, C), rectangle(C).`

In the following experiments, trains problems are formed from five positive and five negative examples. As the experiments show, randomly selected problems like this can be quite hard for an ILP system to solve

The background knowledge for train `t3` in Figure 5.1 is encoded in Prolog as shown in Figure 5.2. The full background knowledge for the trains problem contains similar encodings of 1000 trains. The language bias for the problem allows `f` as a head

```
train(t3).                      roof_open(t3_c1).
has_car(t3, t3_c1).             roof_open(t3_c2).
has_car(t3, t3_c2).             has_load(t3_c1, t3_c1_l1).
short(t3_c1).                   has_load(t3_c2, t3_c2_l1).
short(t3_c2).                   triangle(t3_c1_l1).
two_wheels(t3_c1).              rectangle(t3_c2_l1).
two_wheels(t3_c2).
```

Figure 5.2: A logic program constituting background knowledge for train `t3`.

predicate symbol and 22 possible predicate symbols in the body. It also contains `type` and `direction` declarations for each predicate symbol. The settings used for the trains problem include `max_vars(5)`, `max_clauses(4)`, and `max_body(5)`. The full language bias for the trains problem can be found in Appendix F. The full BK is over 28,000 lines and too long to include. A link to the full BK is provided in in Appendix F

### 5.1.1 Random Trains Problems

The first experiment tests **Q1**, **Q2**, **Q4**, **Q5**, and **Q6** by comparing Popper and ⊥-Popper on random trains problems. The background knowledge, language bias, and settings for each trains problem are as described above.

**Methods** 200 instances of the trains problem were generated. Each problem instance was generated by selecting 5 random trains in the BK as positive examples and 5 different random trains as negative examples. Each instance of the trains problem was passed as input to Popper and each ⊥-Popper version. Each system was given five minutes to solve each trains problem. Systems which timed out were recorded as finding no solution for that problem. The following statistics were recorded after executing each system on each problem: execution time, number of programs generated, whether a solution was found, the number of literals in the solution, and the number of bottom clause variants generated. A solution for any version of ⊥-Popper was considered *non-optimal* if Popper was able to find a solution with less literals.

**Results** The mean results for this experiment are shown in Table 5.1.

From this experiment, it seems the answer to **Q1** is yes. On random trains problems bottom preprocessing can reduce the number of programs generated by Popper on average by a factor of over 1000. This shows that bottom preprocessing can substantially reduce the space of programs a meta-level ILP system explores. Bottom preprocessing has a cost though. **Q2** asks if this reduction in the hypothesis space improves the runtime.

|  | ⊥-**Popper** | **Popper** | ⊥-**Popper**$_+$ | ⊥-**Popper**$_-$ | ⊥-**Popper**$_{VS}$ |
|---|---|---|---|---|---|
| Execution time | **18 ± 3** | 197 ± 9 | 140 ± 9 | 163 ± 10 | 20 ± 3 |
| Programs generated | **7 ± 5** | 8611 ± 393 | 3168 ± 205 | 4382 ± 265 | **7 ± 5** |
| Percent solved | **96** | 45 | 66 | 55 | **96** |
| Percent optimal | 99.5 | **100** | 99.5 | **100** | 99.5 |
| ⊥ variants | 2927 ± 75 | N/A | 1474 ± 46 | 1520 ± 50 | 2918 ± 77 |

Table 5.1: Aggregate results on random trains problems for Popper and versions of ⊥-Popper. The rows are, from top to bottom, mean execution time in seconds per problem, mean number of programs generated per problem, percentage of total problems solved, percentage of problems solved optimally, and mean number of bottom clause variants generated per problem. The error given is standard error.

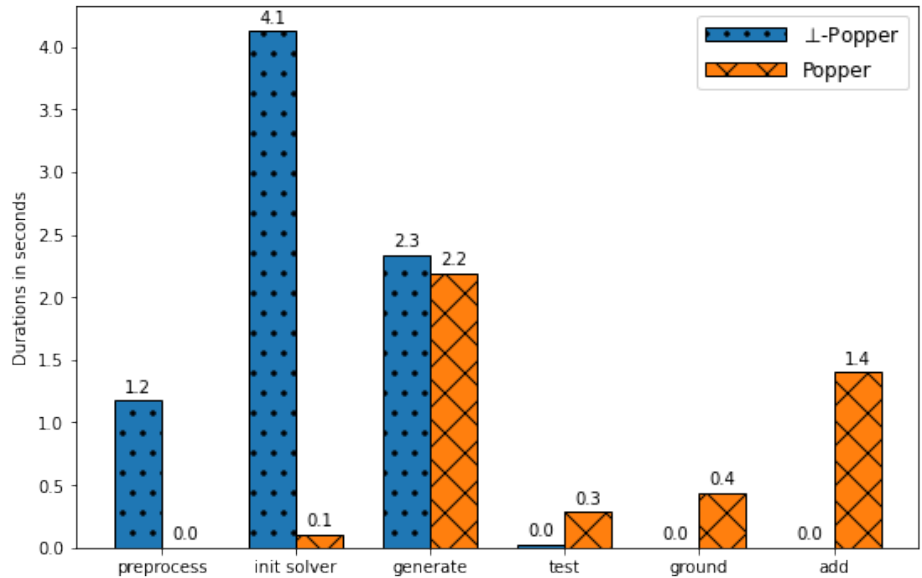

Figure 5.3: Mean execution time spent in each step of finding a solution for random trains problems solvable by Popper in less than 20 seconds.

The random trains experiment shows that the answer to **Q2** is yes. Bottom preprocessing can reduce learning times of a meta-level ILP system on hard problems. ⊥-Popper was 10.8 times faster on average than Popper when solving random trains problems. Perhaps more importantly, ⊥-Popper was able to find a solution for 96 percent of the problems while base Popper could only find a solution for 45 percent.

Bottom preprocessing has an initial cost. It must compute bottom clauses and then ground the additional hypothesis constraints it sends to ASP. Figure 5.3 shows the mean amount of time Popper and ⊥-Popper spent in each step of finding a solution. It only includes the subset of trains problems which Popper could solve in less than 20 seconds. For these problems, the mean execution time for Popper was 4.4 seconds while the mean execution time for ⊥-Popper was 7.6 seconds. The stage 'init solver' involves grounding the initial hypothesis constraints for a problem. This is where most of the overhead from bottom preprocessing occurs. Figure 5.3 shows that for problems with answers that were easy for Popper to find quickly, the overhead of preprocessing seemed to dominate ⊥-Popper's runtime.

This experiment shows that the answer to **Q4** is yes. Both ⊥-Popper$_-$ and ⊥-Popper$_+$ perform better than Popper. They also generate fewer programs. However, it is the combination of both sets of constraints in ⊥-Popper that produces the most significant improvements for the random trains problems. It would seem that both positive and negative bottom preprocessing are useful.

The answer to **Q5** is also yes. The unsoundness of positive example bottom preprocessing is not a significant problem on random trains problems. Only a single non-optimal solution was produced by any version of ⊥-Popper. For one problem, ⊥-Popper, ⊥-Popper$_+$, and ⊥-Popper$_{VS}$ found a solution with six literals while Popper was able to find a solution with five literals. As expected, ⊥-Popper$_-$ was able to find the optimal solution in this case.

For this experiment, the answer to **Q6** is no. Variable splitting did not reduce performance significantly. The structure of the trains problem is such that there are rarely any variables to split. Thus ⊥-Popper$_{VS}$ produced similar bottom clause variants to ⊥-Popper and performed only slightly worse.

### 5.1.2 Irrelevant Background Knowledge

The random trains problems experiment did not answer **Q3**. It did not show anything about how bottom preprocessing affects performance in the presence of irrelevant background knowledge. To determine this, another experiment was run on a single trains problem which was modified to have increasing amounts of irrelevant BK. The starting background knowledge, starting language bias, and settings for each trains problem were the same as those used in the random trains problems experiment.

**Methods** A single set of five positive and five negative examples was used for every problem. The timeout given to each problem was ten minutes. Two different methods were tested for expanding the starting BK and language bias.

**Dyadic predicate symbol**   In the first configuration 100 problems were generated by adding $i$, ranging from 1 to 100, additional irrelevant predicate symbols to the language bias and BK. Thus $L_i$ was $L_{i-1}$ with the following addition:

```
body_pred(has_useless_{i},2).
direction(has_useless_{i},(in, out)).
type(has_useless_{i},(train, car)).
```

$B_i$ was $B_{i-1}$ with the following addition:

```
has_useless_{i}(t1000, t1000_c1).
```

Since the train `t1000` was not one of the trains used in the examples, this additional BK was not relevant to the problem. This is called the *dyadic* configuration because the added predicate symbols have arity 2.

**Monadic predicate symbol**   In the second configuration 200 trains problems were generated in a similar manner to the dyadic configuration. However, the added predicate symbols had arity 1, making this the monadic configuration. $L_i$ was $L_{i-1}$ with the following addition:

```
body_pred(c{i},1).
direction(c{i},(out,)).
type(c{i},(load,)).
```

$B_i$ was $B_{i-1}$ with the following addition:

```
c{i}(t1000_c1_l1).
```

Again since the train `t1000` was not one of the trains used in the examples, this additional BK was not relevant to the problem. `t1000_c1_l1` is a load which can only appear on `t1000`.

**Results**   The effect of the additional predicate symbols on the execution time of Popper and three versions of ⊥-Popper is shown in Figure 5.4 and Figure 5.5.

The answer to **Q2** is yes. Figure 5.4 shows that bottom preprocessing can in some cases significantly improve the scaling of Popper when irrelevant background knowledge is present. However, the scaling is not improved when the additional predicate symbols are monadic, as shown in Figure 5.5

Note that the direction declarations in the monadic configuration, `direction( c{i},(out,))`, have direction `out` for all arguments. Monadic predicate symbols with this form of direction declaration are used in Popper as *constant surrogates*. Constant surrogates are monadic predicate symbols that represent a constant symbol in an LFF hypothesis. They are necessary because ground terms are not allowed in LFF hypotheses. To include a constant such as 0, it is necessary to introduce a constant surrogate, such as `zero/1`.

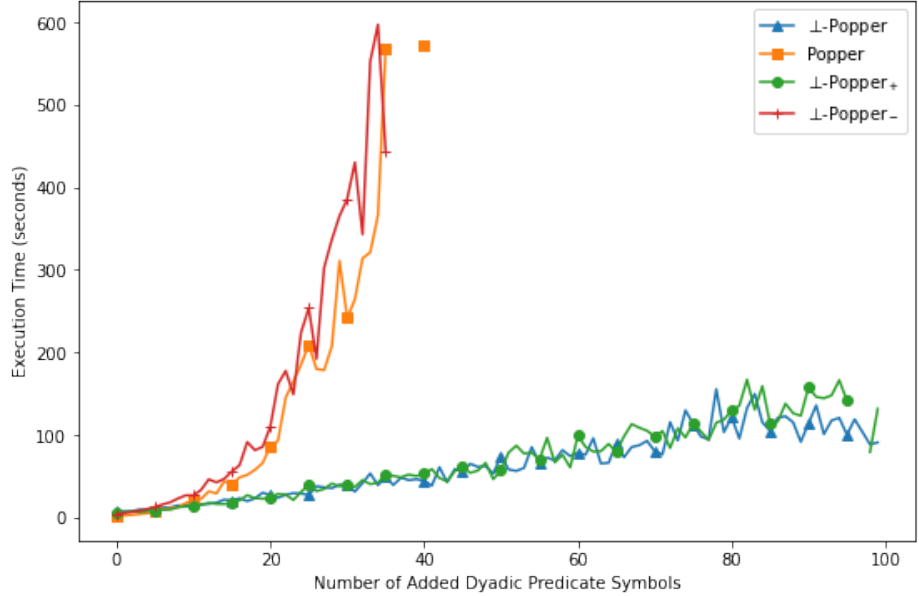

Figure 5.4: Execution time of Popper and ⊥-Popper as irrelevant dyadic predicate symbols are added to the BK of a trains problem.

The difficulty for ⊥-Popper arises when a constant surrogate such as `zero(0)` is a fact that is true in every model of $B$. Since `zero(0)` is always bias consistent, it occurs in the bottom clause of every example relative to $B$. Since the number of bottom clause encodings generated by ⊥-Popper scales with the number of literals in the bottom clause, every constant surrogate introduced reduces performance.

Suppose $l$ is a literal with the dyadic predicate symbol `has_useless_i/2`. In the dyadic configuration, the direction declaration `direction(has_useless_i,(in, out))` suffices to exclude $l$ from all bottom clauses of examples. This is because $l$ only occurs in a bottom clause ⊥ if the first argument of $l$ is instantiated by another literal in ⊥. Given the BK used in the experiment, the first argument of $l$ is always `t1000`. Since `t1000` is not relevant to any of the examples in the problem, `t1000` will never be instantiated by any other literal. Thus the predicate symbol `has_useless_i/2` does not occur in any positive example bottom clause. In that case, bottom preprocessing hypothesis constraints ensure it cannot occur in any solution.

This shows that direction declarations are quite important when using bottom preprocessing. In cases where the language bias can exclude irrelevant literals from a bottom clause, bottom preprocessing can improve scaling. It is also worth noting that the positive example bottom preprocessing seems to do most of the work. ⊥-Popper₋ has the same scaling issues as Popper does in the presence of irrelevant BK.

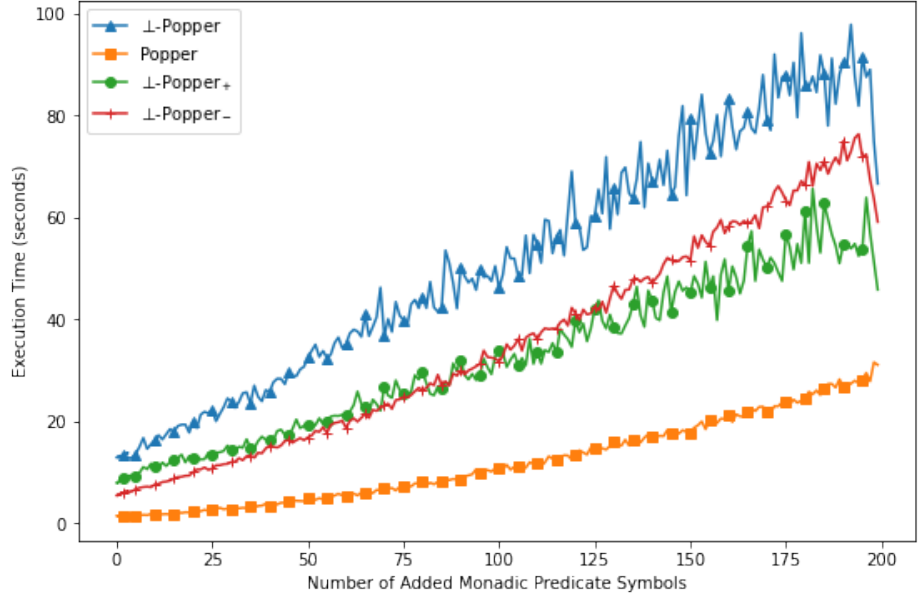

Figure 5.5: Execution time of Popper and ⊥-Popper as irrelevant monadic predicate symbols are added to the BK of a trains problem.

## 5.2 List Manipulation

Chapter 4 described how positive bottom preprocessing uses weaker constraints for recursive hypotheses. This experiment tests whether bottom preprocessing is still effective on problems that require recursion. The original Popper paper used list manipulation problems to evaluate Popper's performance on difficult recursive tasks. The same task set and methodology is used here.

An example of a list manipulation problem is `dropk`. The `dropk` task is to find a relation $f(L_1, k, L_2)$ such that $L_2$ is $L_1$ without its first $k$ elements. Thus `f([3, 2, 1], 1, [2, 1])` is a positive example but `f([3, 2, 1], 1, [3, 1])` is a negative example. A solution to this problem as a logic program is the following:

```
f(A,B,C) :- one(B),tail(A,C).
f(A,B,C) :- tail(A,D),decrement(B,E),f(D,E,C).
```

The nine list manipulation problems used in this experiment are `dropk`, `droplast`, `evens`, `finddup`, `last`, `len`, `member`, `sorted`, `sumlist`. The language bias provided to the tasks includes the following relations: `head/2`, `tail/2`, `geq/2`, `empty/1`, `even/1`, `odd/1`, `one/1`, `zero/1`, `decrement/2`. The following ad-ditional bias was given to some tasks. `droplast` was given `cons/3`. `finddup` was given `element/2`. `len` was given `increment/2`. `sum` was given `sum/2`. Type and direction declarations for all relations were provided. Background knowledge implementing all the listed relations was also provided. The full background knowledge and language bias for each task can be found in Appendix E. All problems were given settings of `max_vars(5)`, `max_clauses(2)`, and `max_body(5)`.

| Task | ⊥-Popper | Popper | ⊥-Popper$_+$ | ⊥-Popper$_-$ | ⊥-Popper$_{VS}$ |
|---|---|---|---|---|---|
| dropk | **17 ± 1.7** | 21 ± 1.5 | 19 ± 1.8 | 40 ± 2.5 | 17 ± 1.6 |
| droplast | 75 ± 6.0 | 24 ± 2.1 | **23 ± 1.8** | 121 ± 8.7 | 405 ± 27.7 |
| evens | 17 ± 0.5 | **14 ± 0.2** | 15 ± 0.5 | 42 ± 1.0 | 17 ± 0.6 |
| finddup | **8 ± 0.5** | 44 ± 1.5 | 8 ± 0.7 | 75 ± 3.6 | 21 ± 1.0 |
| last | 4 ± 0.3 | 5 ± 0.3 | **3 ± 0.2** | 14 ± 0.8 | 5 ± 0.3 |
| len | 4 ± 0.1 | 19 ± 0.6 | **3 ± 0.1** | 34 ± 2.9 | 9 ± 0.4 |
| member | 2 ± 0.0 | 2 ± 0.0 | **1 ± 0.0** | 3 ± 0.2 | 2 ± 0.0 |
| sorted | 86 ± 6.9 | **50 ± 4.4** | 68 ± 6.2 | 114 ± 6.7 | 116 ± 9.2 |
| sumlist | 11 ± 0.7 | 430 ± 12.2 | **8 ± 0.4** | 450 ± 10.0 | 113 ± 7.5 |

Table 5.2: Mean execution time per list manipulation task. Results are for Popper and versions of ⊥-Popper. The error given is standard error.

**Methods**  10 random positive and negative examples were used to generate each problem. 20 such problems per task were then passed as input to each system. The timeout was set to 10 minutes for each problem. The parameter `eval_timeout` was set to .001. This enforced a 1ms timeout on testing hypotheses. Similarly to the trains problems the following statistics were recorded after executing each system on each problem: execution time, number of programs generated, whether a solution was found, the number of literals in the solution, and the number of bottom clause variants generated. Each solution was evaluated by running it on 1000 random positive and 1000 random negative examples and recording its accuracy.

**Results**  The mean execution time for each task on each version is summarized in Table 5.2. The mean number of programs generated for each task on each system is summarized in Table 5.3. Table 5.4 shows the mean number of bottom clause variants generated for each task on each system.

This experiment shows that the answer to **Q1** is again yes. ⊥-Popper explored less programs than Popper on every task. Note that this question differs from the question of whether bottom preprocessing can reduce the initial hypothesis space of a meta-level ILP system. This second question is answered theoretically in Chapter

| Task | ⊥-Popper | Popper | ⊥-Popper$_+$ | ⊥-Popper$_-$ | ⊥-Popper$_{VS}$ |
|---|---|---|---|---|---|
| dropk | 251 ± 31 | 355 ± 31 | 339 ± 37 | 616 ± 38 | **239** ± 29 |
| droplast | **95** ± 13 | 154 ± 15 | 107 ± 14 | 177 ± 18 | 128 ± 18 |
| evens | 46 ± 3 | 438 ± 3 | 144 ± 4 | 455 ± 7 | **43** ± 3 |
| finddup | 163 ± 18 | 1513 ± 41 | 284 ± 32 | 1818 ± 58 | **151** ± 16 |
| last | **73** ± 10 | 313 ± 23 | 84 ± 9 | 484 ± 24 | 88 ± 10 |
| len | 49 ± 4 | 712 ± 21 | 50 ± 3 | 914 ± 63 | **49** ± 3 |
| member | **1** ± 0 | 75 ± 2 | 1 ± 0 | 69 ± 5 | 1 ± 0 |
| sorted | **601** ± 54 | 947 ± 56 | 836 ± 71 | 992 ± 48 | 630 ± 56 |
| sumlist | 31 ± 4 | 4125 ± 40 | 32 ± 3 | 4071 ± 7 | **29** ± 4 |

Table 5.3: Mean number of programs generated per list manipulation task. Results are for Popper and versions of ⊥-Popper. The error given is standard error.

| Task | ⊥-Popper | ⊥-Popper$_+$ | ⊥-Popper$_-$ | ⊥-Popper$_{VS}$ |
|---|---|---|---|---|
| dropk | 340 ± 11 | 157 ± 2 | 183 ± 10 | **413** ± 16 |
| droplast | 3957 ± 42 | 385 ± 5 | 3572 ± 42 | **20149** ± 64 |
| evens | 3498 ± 31 | 1747 ± 20 | 1751 ± 19 | **3724** ± 11 |
| finddup | 811 ± 9 | 412 ± 8 | 399 ± 6 | **3050** ± 28 |
| last | 763 ± 8 | 364 ± 6 | 398 ± 8 | **790** ± 9 |
| len | 604 ± 19 | 310 ± 15 | 294 ± 12 | **1696** ± 58 |
| member | 726 ± 9 | 388 ± 8 | 338 ± 6 | **766** ± 10 |
| sorted | 2924 ± 41 | 1157 ± 31 | 1767 ± 19 | **3830** ± 31 |
| sumlist | 797 ± 36 | 415 ± 28 | 382 ± 18 | **6357** ± 706 |

Table 5.4: Mean number of bottom clause variants generated per list manipulation task. Results are for Popper and versions of ⊥-Popper. The error given is standard error. The **bolded** values in this table are maximums.

3. The experimental results shown here demonstrate that this reduction in the initial hypothesis space translates into ⊥-Popper exploring less programs overall before finding a solution. The results for ⊥-Popper$_-$ show that this is not always the case. ⊥-Popper$_-$ frequently explores more programs than Popper. This phenomenon is explored in more depth in relation to **Q4** below.

The answer to **Q2** is also yes. ⊥-Popper or ⊥-Popper$_+$ demonstrated improved performance over Popper on all but two tasks: `evens` and `sorted`. In this case ⊥-Popper$_+$ was the most consistent performer. It was only significantly slower than Popper on the `sorted` task, taking 1.36x as long on average. However, it was 51.9x faster on `sumlist`, 6.5x faster on `len`, and 5.4x faster on `finddup`.

For this experiment, the answer to **Q4** is no. Positive and negative example bottom preprocessing are not equally effective. In fact, in all cases negative bottom preprocessing is strictly worse on these recursive tasks than no preprocessing at all.

There are several reasons for the poor performance of ⊥-Popper$_-$. First, no re-cursive clauses generalize a negative example bottom clause since the head predicate symbol of any clause in a hypothesis cannot occur in the bottom clause. This means negative bottom preprocessing can only prune recursive hypotheses due to non-recursive clauses.

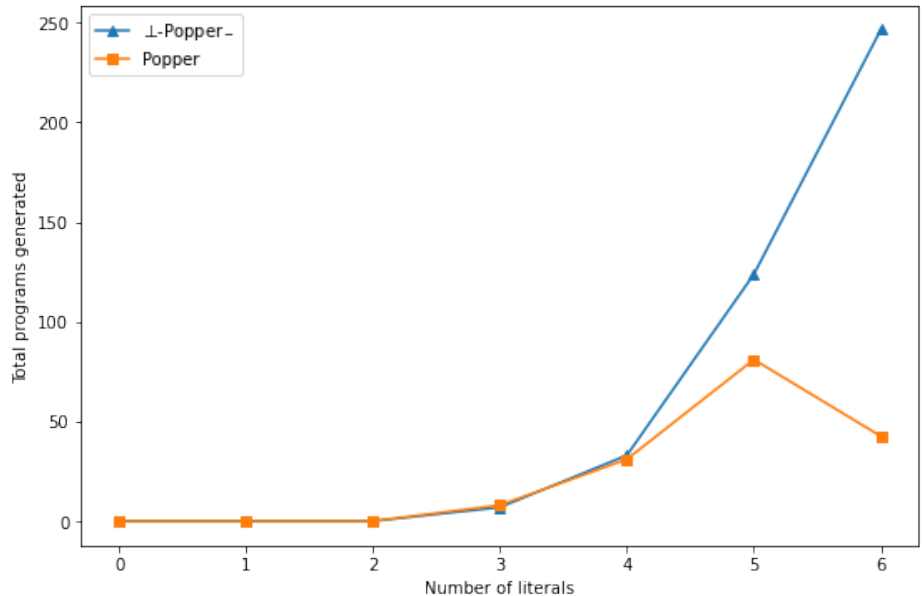

Figure 5.6: Total number of programs with $n$ literals generated by $\bot$-Popper$_-$ and Popper as $n$ increases. These results are for a randomly chosen `dropk` task.

Second, it is possible for hypothesis constraints to prune hypotheses that help Popper learn better constraints. This happens in the cases where $\bot$-Popper$_-$ generates more programs than Popper. Figure 5.6 shows, for the `dropk` task, the total number of programs with $n$ literals generated by $\bot$-Popper$_-$ and Popper as $n$ increases. In this case, negative bottom preprocessing prunes slightly more hypotheses of size 3. Yet it then generates many more programs of larger sizes, presumably because Popper learns better constraints from the programs negative bottom preprocessing never allows it to evaluate. Popper can learn multiple constraints from a failed hypothesis, including both generalization and specialization constraints.

Third, negative example bottom preprocessing can generate hypothesis constraints that increase the time it takes for ASP to generate any program. On one `droplast` problem, $\bot$-Popper$_-$ spends 12 seconds preprocessing and grounding constraints while Popper only spends 0.7 seconds. This time investment doesn't pay off though as $\bot$-Popper$_-$ spends 130 seconds generating a larger number of hypotheses then Popper. Popper only takes 16 seconds to generate hypotheses in this case. The extra constraints make the CSP search harder without actually causing Popper to generate

| Task | $\perp$-**Popper** | **Popper** | $\perp$-**Popper**$_+$ | $\perp$-**Popper**$_-$ | $\perp$-**Popper**$_{VS}$ |
|---|---|---|---|---|---|
| dropk    | **99 ± 3**  | 100 ± 0 | 100 ± 2   | **99 ± 3**  | **99 ± 2**  |
| droplast | 100 ± 0     | 100 ± 0 | 100 ± 0   | 100 ± 0     | **90 ± 21** |
| evens    | 100 ± 0     | 100 ± 0 | 100 ± 0   | 100 ± 0     | 100 ± 0     |
| finddup  | 99 ± 2      | 99 ± 2  | 99 ± 2    | 99 ± 2      | 99 ± 1      |
| last     | 100 ± 0     | 100 ± 0 | 100 ± 0   | 100 ± 0     | 100 ± 0     |
| len      | 100 ± 0     | 100 ± 0 | 100 ± 0   | 100 ± 0     | 100 ± 0     |
| member   | 100 ± 0     | 100 ± 0 | 100 ± 0   | 100 ± 0     | 100 ± 0     |
| sorted   | 97 ± 8      | 97 ± 8  | **98 ± 7**| **98 ± 7**  | **99 ± 5**  |
| sumlist  | 100 ± 0     | 100 ± 0 | 100 ± 0   | 100 ± 0     | 100 ± 0     |

Table 5.5: Mean accuracy for each list manipulation task. The error given is standard deviation. Bolded text highlights instances where $\perp$-Popper and Popper differed.

fewer hypotheses.

Thus it could be better for $\perp$-Popper to simply not perform negative example bottom preprocessing when the problem allows recursive solutions.

The answer to **Q5** is yes. All versions of $\perp$-Popper were able to find a solution for all problems except two instances in which $\perp$-Popper$_{VS}$ timed out on `droplast`. Only two non-optimal solutions were found for 180 problems. In one case, $\perp$-Popper$_-$ found a non-optimal solution. This could indicate a bug in the code. There were no cases in which variable splitting was necessary to find an optimal solution. Figure 5.5 compares the accuracy of the different versions on the list manipulation tasks. None of the differences in the table are stastically significant. The fact that any differences exist on `dropk` and `droplast` can be explained by the two instances of non-optimality and the two timeouts described above. The differences on `sorted` are assumed to be random chance.

The answer to **Q6** is a weak yes. $\perp$-Popper$_{VS}$ was always slower than $\perp$-Popper. Sometimes it was significantly slower. It achieved no benefit since it never found an optimal solution when $\perp$-Popper$_+$ did not. Table 5.4 shows that $\perp$-Popper$_{VS}$ generated many more bottom clause variants than $\perp$-Popper$_+$ in all cases. For instance, on the `droplast` task $\perp$-Popper$_{VS}$ produced 52.3 times more variants and performed 18 times slower than $\perp$-Popper$_+$. $\perp$-Popper$_{VS}$ also had lower accuracy. Future work could explore other datasets designed specifically to require variable splitting.

# Chapter 6

# Related Work

This chapter provides a summary of research on inductive logic programming related to bottom preprocessing. It serves as a short literature review of inverse entailment and meta-level ILP systems. Rather than attempt comprehensiveness, it highlights the most relevant research for bottom preprocessing. This chapter also discusses the similarities and differences between existing research and the generation of hypothesis constraints through preprocessing. This is intended to assist the reader in assessing the novelty of this dissertation's contributions.

## 6.1 Inverse Entailment

The inverse entailment approach to ILP was introduced in a 1995 paper by Stephen Muggleton [20]. This paper also introduced the bottom clause and the first inverse entailment implementation, Progol. Inverse entailment and Progol have been highly influential in ILP. Further systems which extend inverse entailment include Aleph [27], HAIL [26], CF-Induction [11], Imparo [12], and ATOM [1].

In Definition 24, inverse entailment was defined as deriving a clause $C$ from an example $e$, given background knowledge $B$, if and only if $C$ subsumes $\perp_B(e)$. Different implementations of inverse entailment take different approaches to finding the clause $C$ given $\perp_B(e)$. Progol makes use of the fact that $\perp_B(e)$ imposes a greatest lower bound on a lattice formed by the subsumption relation. This lattice has the empty clause $\square$ at its top. Given a problem input $(B, M, E+, E-)$ containing background knowledge, mode declarations, positive examples and negative examples respectively, Progol proceeds as follows:

1. Initialize the current hypothesis $H$ to an empty set of clauses.

2. Choose a positive example $e \in E^+$ which is not covered by $H$. If all positive examples are covered, return $H$.

3. Compute $\perp_{B,M}(e)$, the mode declaration consistent bottom clause of $e$, using the algorithm described in chapter 4.

4. Use A* search to find the optimal clause $C_e$ in the lattice between $\square$ and $\perp_{B,M}(e)$. This search starts with a clause $C_e = \square$. It uses a specialization operator, $\rho$, to produce and evaluate candidate specializations of $C_e$. If it finds a better candidate, the candidate becomes $C_e$. The A* heuristic used can take many forms. The original choice was Occam *compression* which Aleph defines as "P - N - L + 1, where P, N are the number of positive and negative examples covered by the clause, and L the number of literals in the clause [27]."

5. After finding the optimal clause $C_e$ for some $e$, add $C_e$ to $H$ and returns to step 2.

As discussed in chapter 3, Yamamoto showed that inverse entailment is not complete for entailment [30]. In the same paper, he showed that inverse entailment is complete for relative subsumption. Muggleton attempted to resolve the incompleteness of inverse entailment by expanding the computation of the bottom clause for inverse entailment but, as he noted, this made inverse entailment unsound [21].

Yamamoto later defined a *bridge theory* $F$, given background knowledge $B$ and example $e$, as any ground clausal theory such that $B \cup \neg e \models F$ [31]. The bottom clause of $e$ with respect to B is one of many possible bridge theories. Several inverse entailment based systems introduce bridge theories that extend inverse entailment to cover new cases. HAIL [26], for instance, uses a bridge theory called *kernel sets*. Kernel sets enable clausal theories to be learned from a single example using inverse entailment. HAIL extends the possible hypotheses that can be learned by inverse entailment outside the bounds of Plotkin's relative subsumption.CF-Induction [11] presents a bridge theory called *characteristic clauses* which extends inverse entailment to full clausal theories.

The bottom preprocessing approach described in this dissertation makes use of the bottom clause as originally described with Progol. Alternative bridge theories could possibly be used in future work to expand the hypotheses positive bottom preprocessing allows.

The ILP system ATOM [1] introduces the use of a solver to inverse entailment. ATOM encodes the clausal search in Progol as a constraint satisfaction problem. Similarly to bottom preprocessing, the encoding ATOM uses ensures that every clause in a hypothesis generalizes the bottom clause of a positive example. Since bottom preprocessing generates hypothesis constraints in the hypothesis language of a target ILP system, ATOM's encoding differs significantly from $\bot$-Popper's. ATOM also searches for clauses that generalize a single example at a time while bottom preprocessing constrains entire hypotheses. ATOM also does not make use of negative examples in its initial constraints. Instead, when it finds a clause $C$ which entails a negative example, it generates pruning constraints that eliminate generalizations of $C$ from its search. This approach to handling negative examples is more similar to Popper's learning from failures than bottom preprocessing.

## 6.2 Meta-level ILP

Meta-level ILP systems generally differ from inverse entailment based systems by evaluating entire hypotheses rather than iteratively finding individual clauses. Since Popper is a meta-level ILP system and $\bot$-Popper is an extension of Popper, this section begins with an explanation of how Popper works. Popper implements the learning from failures approach which this dissertation's problem setting is based on. Popper encodes an LFF problem as a constraint satisfaction problem on the syntax of logic programming, specifically as an ASP program. Popper uses a generate-test-constrain loop. In each loop:

1. The ASP solver **generates** a hypothesis program.

2. Popper **tests** the hypothesis against the examples.

3. If the program fails, Popper **constrains** future hypotheses in the ASP solver.

Popper can generates three types of constraints from a failing hypothesis $H$:

- Popper generates a *generalization* constraint when $H$ entails a negative example. In this case, no generalization of $H$ can be a consistent solution.

- Popper generates a *specialization* constraint when $H$ does not entail a positive example. In this case, no specialization of $H$ can be a complete solution.

- Popper generates an *elimination* constraint when $H$ does not entail any positive examples. Elimination constraints disallow $H$ from appearing as a subset of the clauses in any hypothesis that cannot self-resolve.

Bottom preprocessing can be viewed as taking Popper's approach of generating constraints from a hypothesis and extending it to generating constraints from the bottom clauses of examples. Note that specialization constraints generated from bottom clauses would be sound. Any specialization of the bottom clause of an example $e$ does not even entail $e$ itself. These specialization constraints were not considered for bottom preprocessing as they are too specific to be useful in most cases.

ASPAL [3] and ILASP [15] are also meta-level systems which encode ILP problems using ASP constraints. These systems differ from Popper and $\bot$-Popper in that they learn ASP programs and expect BK written in ASP. They can learn non-monotonic programs from non-monotonic BK. This means hypothesis constraints can be provided directly in the BK. However, they struggle with infinite domains [7]. Future work could explore whether bottom preprocessing is also effective when used to generate initial constraints for these systems.

FastLAS [14] is the system which employs the most similar approach to bottom preprocessing. FastLAS uses the Learning from Answer Sets setting introduced with ILASP. It takes a *scoring function* which ranks hypotheses. It produces an optimal solution according to this scoring function. The first phase of FastLAS computes the *characterization* of each positive and negative example for a LAS problem $T$. It then sucessively refines three hypothesis spaces from these characterizations called respectively the reduced, generalized and optimized characteristic hypothesis spaces of $T$. The solving step of the FastLAS algorithm can be performed by passing the optimised characteristic hypothesis space and the problem specification $T$ to ILASP.

Thus FastLAS can perhaps be viewed as an implementation of hypothesis constraint preprocessing. It generates hypothesis constraints from bridge theories of examples and then uses those constraints to prune the hypothesis space of an ILP problem. However, the theoretical framework and algorithm of FastLAS differ substantially from bottom preprocessing. FastLAS generates ASP programs not definite programs. Characterizations are computed differently than bottom clauses. The constraints enforced on the hypothesis space are entirely different from the subsumption based generalization constraints of bottom preprocessing. The FastLAS algorithm is also deeply tied to the optimization of solutions based on the scoring function it takes as input. Finally FastLAS suffers from the same limitations mentioned for ILASP.

# Chapter 7

# Conclusions

This dissertation introduced *bottom preprocessing*, a method for pruning the initial hypothesis space of an ILP problem. Bottom preprocessing generates a set of hypothesis constraints from the bottom clauses of a problem's examples. The constraints it generates are based on two ideas. Any hypothesis $H^-$ that generalizes the bottom clause of a negative example also entails that negative example. Thus $H^-$ is inconsistent and can be pruned from the hypothesis space. Any hypothesis $H^+$ that does not generalize the bottom clauses of all positive examples is unlikely to entail the examples. Pruning $H^+$ is not sound but it was shown that doing so works well in practice.

Bottom preprocessing is implemented in $\bot$-Popper, an extension to the meta-level ILP system Popper [7]. $\bot$-Popper encodes the ideas above as constraints in ASP. It incorporates Progol's bottom clause construction algorithm to find bottom clauses for each example in an ILP problem. It then encodes many *bottom clause variants* to encode the subsumption check between a hypothesis and a bottom clause.

Experiments were run on the Michalski trains problem and list manipulation programming problems. It was shown that bottom preprocessing can significantly reduce the initial hypothesis space of an ILP problem. Bottom preprocessing can also reduce learning times of a meta-level ILP system on difficult problems. In some cases, bottom preprocessing can enable an ILP system to efficiently handle large background knowledge with many irrelevant predicate symbols.

## 7.1 Limitations

This section summarizes several limitations of bottom preprocessing that could be addressed in future work.

**Soundness** Chapter 3 showed that sound hypothesis constraints cannot be generated from positive examples using inverse entailment. The experiments performed with the trains problem and list manipulation showed that this unsoundness may not be a problem in practice. However, unsoundness compromises the optimality guarantee of an ILP system like Popper. Future work could try other bridge theories for inverse entailment such as kernel sets [26] or characteristic clauses [11].

Chapter 3 also showed that a complete generalization relation is necessary for positive bottom preprocessing to generate subsumption-sound constraints. Chapter 4 used variable splitting to expand its generalization relation toward completeness. Chapter 5 showed that variable splitting could be very expensive when it causes bottom preprocessing to generate many extra bottom clause variants. One approach to resolving this issue could be to find a generalization relation that is complete but not sound. As mentioned in chapter 3, a generalization relation which holds for all possible pairs of clauses would satisfy this requirement. It would not prune any hypotheses though. A complete, unsound generalization relation with a tighter bound might prove quite useful for positive bottom preprocessing.

**Initial cost** The experiments with random trains problems showed that bottom preprocessing has an initial cost on easier problems that it doesn't always overcome. One approach to addressing this is to perform bottom preprocessing lazily rather than up front. Popper, for instance, searches hypotheses by incrementally increasing the number of literals allowed in the hypothesis. Intuitively, it is not necessary to compute the bottom clause to its maximum variable depth in cases where the number of possible literals in the hypothesis is small. Further, there might be some subset of bottom clause variants that are irrelevant for a given number of literals in the hypothesis. Lazy generation of the bottom clause and bottom clause variant encodings as the search space expands could amortize some of the up front cost of bottom preprocessing.

**Implementation** Chapter 4 noted that encoding all variants of the variable split bottom clause may not ensure completeness of the generalization relation $\perp$-Popper uses. It was also not clear that this approach was more efficient than alternatives. Since straightforward subsumption checking algorithms exist [8], it should be possible to implement one of these algorithms into ASP. It would then be possible to compare the performance of $\perp$-Popper with a different subsumption check to the current version of $\perp$-Popper. It is possible that this subsumption check could be implemented directly as hypothesis constraints instead of a separate program which encodes bottom clause variants. This approach is appealing because it might enable interactions between the constraints Popper learns and the subsumption check of bottom preprocessing.

**Constants** The experiment which added monadic predicate symbols to the language bias of a trains problem showed that $\bot$-Popper does not scale well when constant surrogates are present. Since a non-bias consistent bottom clause is ground before being lifted, it can be interpreted as containing the full set of ground terms which are relevant to an example. Future research could explore whether this could be used to remove the restriction that the hypothesis contain no ground terms in the LFF problem setting. Recursion and the incompleteness of inverse entailment might make this set of ground terms incomplete. It is still a potentially interesting avenue of investigation. Enabling Popper to use ground terms in hypotheses would have many benefits, one of which would be no longer requiring constant surrogates in the BK of an LFF problem.

## 7.2 Future Work

There are other directions future work with bottom preprocessing could take besides overcoming the current limitations of $\bot$-Popper.

**Predicate Invention** A recent paper extends Popper with *predicate invention*. *Predicate invention* is the ability of an ILP system to invent new predicate symbols that can be used in a hypothesis. Predicate invention can enable an ILP system to expand its own BK during solving [6]. However, predicate invention is difficult for any inverse entailment system because any predicate symbols the system invents will not exist in the bottom clauses of the examples. Bottom preprocessing inherits this limitation. The current implementation of $\bot$-Popper simply disables bottom preprocessing constraints for problems which use predicate invention. Further work could explore a more nuanced approach to integrating predicate invention with bottom preprocessing.

**Other preprocessing approaches** Bottom clauses are not the only information that can be extracted from the examples and background knowledge of an ILP problem. GOLEM [23], one of the earliest ILP systems, used *relative least general generalization* (rlgg) to find solutions to ILP problems. The rlgg is a single clause that covers a set of examples. ATOM showed that constraints can be generated on hypotheses if predicate symbols are marked as functional, i.e. the truth value is not important [1]. Future work could explore these and potentially other creative ways to efficiently generate initial constraints for meta-level ILP systems.

# Appendix A

# Mode declaration code

The following Prolog code can generate a list of mode declarations from an LFF language bias. The definitions of `head_preds` and `body_preds` can be found in Appendix ??.

```prolog
1   modes(L) :-
2       head_preds(Heads),
3       body_preds(Bodies),
4       maplist(make_mode(head), Heads, HeadModes),
5       maplist(make_mode(body), Bodies, BodyModes),
6       L1 = [HeadModes, BodyModes],
7       flatten(L1, L).
8
9   make_mode(Slot, (PredName/_, _, Meta), FinalMode) :-
10      make_type_dirs(Meta, [], TypeDirAtoms),
11      FinalPred=..[PredName|TypeDirAtoms],
12      (Slot=head -> FinalMode=(:- modeh(1, FinalPred));
13      Slot=body -> FinalMode=(:- modeb(*, FinalPred))).
14
15  make_type_dirs([], LIn, LOut) :- reverse(LIn, LOut).
16  make_type_dirs([(Type, Direction)|Meta], LIn, Atoms) :-
17      (Direction=in, DirChar=(+);
18      Direction=out, DirChar=(-)),
19      atom_concat(DirChar, Type, Atom),
20      make_type_dirs(Meta, [Atom | LIn], Atoms).
```

# Appendix B

# `bc_prog` code

```prolog
:- dynamic
           type/2,
           neg/1,
           pos/1,
           head_pred/2,
           body_pred/2,
           direction/2.

%% Bias handling
head_preds(Preds) :-
    preds(head, Preds).
body_preds(Preds) :-
    preds(body, Preds).

preds(Slot, Preds) :-
    findall((Name/Arity, Term, Variables),
    pred_ext(Name, Arity, Slot, Term, Variables), Preds).

pred_ext(Name, Arity, Slot, Term, Meta) :-
                          find_pred(Name, Arity, Slot, Types,
                          ↪ Directions),
                          pick_vars(Arity, Vars),
                          Term =.. [Name | Vars],
                          zip(Types, Directions, Meta).

pred(Name, body, Arity) :-
    body_pred(Name, Arity).
pred(Name, head, Arity) :-
    head_pred(Name, Arity).

find_pred(Name, Arity, Slot, Types, Directions) :-
    resolve_pred(Name, Arity, Slot, Directions),
    resolve_type(Name, Arity, Types).

resolve_type(Name, _, Types) :-
    type(Name, TypesT),
    list_from_tuple(TypesT, Types), !.
resolve_type(_, Arity, Types) :-
    none_types(Arity, Types).
```

```prolog
resolve_pred(Name, Arity, Slot, Directions) :-
    pred(Name, Slot, Arity),
    direction(Name, DirectionsT),
    list_from_tuple(DirectionsT, Directions).

none_types(0, []).
none_types(I, [none | L]) :-
    Idec is I - 1,
    none_types(Idec, L), !.

pick_vars(Arity, L) :-
    length(L, Arity). %%Maybe these should be the same variables?

tuple_from_list([A], A) :- !.
tuple_from_list([A|T], (A, T1)) :-
    !, tuple_from_list(T, T1).

list_from_tuple(B, [B]) :-
    var(B), !.
list_from_tuple((A, B), [A|L]) :-
    !, list_from_tuple(B, L).
list_from_tuple(B, [B]).

zip([], [], []).
zip([A|As], [B|Bs], [(A,B)|T]) :- zip(As, Bs, T).

%% Bottom clauses
example(positive, X) :-
    pos(X).
example(negative, X) :-
    neg(X).

print_neg_bottom(X, B) :-
    neg_bottom(X, B), print(B).
neg_bottom(X, B) :-
    example_bottom(negative, X, B).

print_pos_bottom(X, B) :-
    pos_bottom(X, B), print(B).
pos_bottom(X, B) :-
    example_bottom(positive, X, B).

example_bottom(Type, X, B) :-
    example(Type, X),
    bottom(X, B).

bottom(Clause, Bottom) :-
    split_clause(Clause, Head, BodyList),
    assert_all(BodyList),
    find_bottom(Head, Bottom),
    retract_all(BodyList).

assert_all([]).
```

```prolog
 93  assert_all([H | T]) :-
 94      assertz(H), !, assert_all(T).
 95
 96  retract_all([]).
 97  retract_all([H | T]) :-
 98      retract(H), !, retract_all(T).
 99
100  find_bottom(Head, Bottom) :-
101      b_setval(variable_index_counter, 0),
102      find_head(Head, TermMapHead, BottomHead, InTerms1),
103      find_body(TermMapHead, InTerms1, BottomBody),
104      Bottom = (BottomHead :- BottomBody), !.
105
106  find_head(Term, TermMap, Mode, InTerms) :-
107      head_preds(L),
108      empty_assoc(TermMapIn),
109      find_head_r(Term, L, TermMapIn, TermMap, Mode, InTerms).
110
111  find_head_r(Term, [(_, ModeTerm, Meta)|_], TermMapIn, TermMap,
     ↪ FinalTerm, InTerms) :-
112      unifiable(ModeTerm, Term, Substitution),
113      !,
114      update_mode_term(ModeTerm, head, Meta, Substitution, TermMapIn,
         ↪ TermMap, FinalTerm, InTerms).
115
116  find_head_r(Term, [_|T], TermMapIn, TermMap, FinalTerm, Substitution)
     ↪ :-
117      find_head_r(Term, T, TermMapIn, TermMap, FinalTerm,
         ↪ Substitution).
118
119  find_body(TermMap, InTerms, BottomBody) :-
120      body_preds(L),
121      find_body_loop_depth(L, 0, [], TermMap, InTerms, BottomBodyL),
122      tuple_from_list(BottomBodyL, BottomBody).
123
124  find_body_loop_depth(_, Depth, BottomBodyIn, _, _, BottomBodyIn) :-
125      max_variable_depth(Depth), !.
126
127  find_body_loop_depth(L, Depth, BottomBodyIn, TermMapIn, InTermsIn,
     ↪ BottomBody) :-
128      find_body_loop_modes(L, BottomBodyIn, TermMapIn, InTermsIn,
         ↪ InTermsIn, TermMapOut, InTermsOut, BottomBodyOut),
129      NewDepth is Depth+1,
130      find_body_loop_depth(L, NewDepth, BottomBodyOut, TermMapOut,
         ↪ InTermsOut, BottomBody).
131
132  find_body_loop_modes([], BottomBodyIn, TermMapIn, _, TempInTerms,
     ↪ TermMapIn, TempInTerms, BottomBodyIn).
133  find_body_loop_modes([(_, BodyTerm, Meta)|T], BottomBodyIn,
     ↪ TermMapIn, InTermsIn, TempInTerms, TermMap, InTerms, BottomBody)
     ↪ :-
134      find_body_term(BodyTerm, Meta, TermMapIn, InTermsIn, TermMapOut,
         ↪ InTermsOut, BottomBodyOut),
135      ord_union(BottomBodyOut, BottomBodyIn, BottomBodyNext),
```

```prolog
136        ord_union(TempInTerms, InTermsOut, NextTempInTerms),
137        find_body_loop_modes(T, BottomBodyNext, TermMapOut, InTermsIn,
           ↪ NextTempInTerms, TermMap, InTerms, BottomBody).
138
139 find_body_term(BodyTerm, Meta, TermMapIn, InTermsIn, TermMap,
    ↪ InTerms, BottomBody) :-
140        in_variables(BodyTerm, Meta, InVars),
141        find_body_substitutions(BodyTerm, InVars, InTermsIn, InTermsIn,
           ↪ [], BodySubs),
142        make_body_substitutions(BodyTerm, BodySubs, Meta, TermMapIn,
           ↪ InTermsIn, TermMap, InTerms, BottomBody).
143
144 in_variables(Term, Meta, InVars) :-
145        term_variables(Term, Variables),
146        zip(Variables, Meta, VarsWithMeta),
147        include(direction_match(in), VarsWithMeta, L1),
148        maplist(extract_var_and_type, L1, InVars).
149
150 extract_var_and_type((Var, (Type, _)), Var-Type).
151
152 find_body_substitutions(_, _, _, [], _, []).
153 find_body_substitutions(BodyTerm, [], _, _, Substitution, [BodySub])
    ↪ :-
154        substitute(Substitution, BodyTerm, BodySub).
155 find_body_substitutions(BodyTerm, [H-Type|T], InTerms,
    ↪ [Term-Type|Terms], Substitution, BodySubs) :-
156        find_body_substitutions(BodyTerm, T, InTerms, InTerms,
           ↪ [H=Term|Substitution], BodySubs1),
157        find_body_substitutions(BodyTerm, [H-Type|T], InTerms, Terms,
           ↪ Substitution, BodySubs2),
158        append(BodySubs1, BodySubs2, BodySubs).
159 find_body_substitutions(BodyTerm, L, InTerms, [_|Terms],
    ↪ Substitution, BodySubs) :-
160         find_body_substitutions(BodyTerm, L, InTerms, Terms,
            ↪ Substitution, BodySubs).
161
162 make_body_substitutions(BodyTerm, BodySubs, Meta, TermMapIn,
    ↪ InTermsIn, TermMap, InTerms, BottomBody) :-
163        maplist(findall_reorder, BodySubs, L1),
164        flatten(L1, L2),
165        maplist(unifiable(BodyTerm), L2, Substitutions),
166        update_body_terms(BodyTerm, Substitutions, Meta, TermMapIn,
           ↪ InTermsIn, [], TermMap, InTerms, BottomBodyTerms),
167        list_to_ord_set(BottomBodyTerms, BottomBody).
168
169 findall_reorder(X, L) :- findall(X, X, L).
170
171 update_body_terms(_, [], _, TermMapIn, InTermsIn, BottomBodyIn,
    ↪ TermMapIn, InTermsIn, BottomBodyIn).
172 update_body_terms(BodyTerm, [Substitution|Substitutions], Meta,
    ↪ TermMapIn, InTermsIn, BottomBodyIn, TermMap, InTerms, BottomBody)
    ↪ :-
173        update_mode_term(BodyTerm, body, Meta, Substitution, TermMapIn,
           ↪ TermMapOut, BottomBodyTerm, InTermsOut),
```

```prolog
        ord_union(InTermsIn, InTermsOut, NextInTerms),
        update_body_terms(BodyTerm, Substitutions, Meta, TermMapOut,
            NextInTerms, [BottomBodyTerm|BottomBodyIn], TermMap, InTerms,
            BottomBody).

update_mode_term(ModeTerm, Slot, Meta, Substitution, TermMapIn,
    TermMap, FinalTerm, InTerms) :-
        match_sub_with_meta(ModeTerm, Substitution, Meta, SubWithMeta),
        filter_subs(SubWithMeta, Slot, InTerms),
        replace_mode_term(ModeTerm, SubWithMeta, TermMapIn, TermMap,
            FinalTerm).

filter_subs(SubWithMeta, Slot, InTerms) :-
    (Slot=head -> Direction=in;
    Slot=body -> Direction=out),
    include(direction_match(Direction), SubWithMeta, L1),
    maplist(extract_term_and_type, L1, InTermsL),
    list_to_ord_set(InTermsL, InTerms).

match_sub_with_meta(Term, Substitution, Meta, SubWithMeta) :-
    term_variables(Term, Variables),
    zip(Variables, Meta, VarsWithMeta),
    maplist(find_term(Substitution), VarsWithMeta, SubWithMeta).

find_term([], _, _) :-
    throw(bad_substitution).
find_term([(Var1=Term)|_], (Var2, Meta), ((Var1=Term), Meta)) :- Var1
    == Var2, !.
find_term([_|T], Pair, Result) :-
    find_term(T, Pair, Result).

extract_term_and_type(((_=Term), (Type, _)), Term-Type).
direction_match(Direction, (_, (_, Direction))).

replace_mode_term(ModeTerm, SubWithMeta, TermMapIn, TermMap,
    FinalTerm) :-
    swap_subs(SubWithMeta, TermMapIn, TermMap, NewSubstitution), !,
    substitute(NewSubstitution, ModeTerm, FinalTerm), !.

swap_subs([], TermMap, TermMap, []).
%% TODO: Do we need to handle the case when the direction is # or
    insert none somehow?
swap_subs([((Var1=Term), (_, none))| T], TermMapIn, TermMap,
    [(Var1=Term)|T2]) :-
    !, swap_subs(T, TermMapIn, TermMap, T2).
swap_subs([(Var1=Term, _)| T], TermMapIn, TermMap,
    [(Var1=NewTerm)|T2]) :-
    get_assoc(Term, TermMapIn, NewTerm), !,
    swap_subs(T, TermMapIn, TermMap, T2).
swap_subs([(Var1=Term, _)| T], TermMapIn, TermMap,
    [(Var1=NewTerm)|T2]) :-
    b_getval(variable_index_counter, N),
    NewTerm = '$VAR'(N),
    N1 is N + 1,
```

```prolog
218         b_setval(variable_index_counter, N1),
219         put_assoc(Term, TermMapIn, NewTerm, TermMapOut),
220         swap_subs(T, TermMapOut, TermMap, T2).
221
222     substitute(_, A, A) :- ground(A), !.
223     substitute([V=S|_], A, S) :- var(A), A == V, !.
224     substitute([_|T], A, B) :- var(A), substitute(T, A, B), !.
225     substitute([], A, A) :- var(A), !.
226     substitute(L, A, B) :- compound(A), A=..[F|T], maplist(substitute(L),
        ↪ T, T2), B=..[F|T2].
227
228     split_clause((Head :- Body), Head, L) :- list_from_tuple(Body, L), !.
229     split_clause(Head, Head, []).
230
231     %% Discovering chains within the bottom clause
232     find_chains((_ :- Body), Result) :-
233         list_from_tuple(Body, BodyLiterals),
234         empty_assoc(VariableAssocs),
235         discover_chains_r(BodyLiterals, VariableAssocs, Result).
236
237     discover_chains_r([], A, A).
238     discover_chains_r([Head | Tail], InAssocs, OutAssocs) :-
239         add_variable_chains(Head, InAssocs, TempAssocs),
240         discover_chains_r(Tail, TempAssocs, OutAssocs).
241
242     add_variable_chains(Term, InAssocs, OutAssocs) :-
243         zip_vars_with_direction(Term, _, ZippedVars),
244         direction_vars(ZippedVars, in, InVars),
245         direction_vars(ZippedVars, out, OutVars),
246         update_chain_assocs(InVars, OutVars, InAssocs, OutAssocs).
247
248     zip_vars_with_direction(Term, Pred, ZippedVars) :-
249         Term =.. [Pred | Args],
250         length(Args, N),
251         direction(Pred, Directions),
252         list_from_tuple(Directions, DirectionL),
253         length(DirectionL, N), !,
254         zip(Args, DirectionL, ZippedVars).
255
256     update_chain_assocs([], _, In, In).
257     update_chain_assocs(InVars, OutVars, In, Out) :-
258         maplist(extract_var_index, InVars, InVarsL),
259         tuple_from_list(InVarsL, InVarsTuple),
260         update_chain_assocs_r(InVarsTuple, OutVars, In, Out).
261
262     update_chain_assocs_r(_, [], In, In).
263     update_chain_assocs_r(InVarsTuple, ['$VAR'(OutVar) | Tail], InAssocs,
        ↪ OutAssocs) :-
264         find_chain_assocs(OutVar, InAssocs, ChainList),
265         ord_add_element(ChainList, InVarsTuple, NewChainList),
266         put_assoc(OutVar, InAssocs, NewChainList, TempAssocs),
267         update_chain_assocs_r(InVarsTuple, Tail, TempAssocs, OutAssocs).
268
269     extract_var_index('$VAR'(N), N).
```

```prolog
find_chain_assocs(OutVar, InAssocs, Out) :-
    get_assoc(OutVar, InAssocs, Out), !.
find_chain_assocs(_, _, []).

direction_vars(L, Dir, Out) :-
    include(has_direction(Dir), L, Pairs),
    maplist(extract_var, Pairs, Out).

has_direction(Dir, (_, Dir)).
extract_var((V, _), V).

%% Making variables unique and adding equality literals for
%  completeness in positive example bottom clauses

complete_bottom_clause((H :- B), H :- B1, EqualityPairs) :-
    max_var_number((H :- B), 0, N),
    N1 is N + 1,
    b_setval(variable_index_counter, N1),
    empty_assoc(InMap),
    list_from_tuple(B, BodyLiterals),
    make_vars_unique(BodyLiterals, InMap, NewBodyLiterals, OutMap),
    duplicate_literals(NewBodyLiterals, OutMap, L1),
    tuple_from_list(L1, B1),
    assoc_to_list(OutMap, EqualityList),
    include(not_empty_list, EqualityList, EqualityListFiltered),
    maplist(pair_to_tuple, EqualityListFiltered, EqualityPairs), !.

not_empty_list(_-T) :-
    length(T, N),
    N > 0.

make_vars_unique([], InMap, _, InMap).
make_vars_unique([T1 | Tail], InMap, [T2 | Tail2], OutMap) :-
    !,
    make_vars_unique(T1, InMap, T2, OutMap1),
    make_vars_unique(Tail, OutMap1, Tail2, OutMap).
make_vars_unique(Term, InMap, OutTerm, OutMap) :-
    compound(Term),
    zip_vars_with_direction(Term, Pred, ZippedVars),
    map_vars_to_new_variables(ZippedVars, InMap, OutVs, OutMap),
    OutTerm =.. [Pred | OutVs].

% TODO Need to add new literals when the in-vars are in InMap
map_vars_to_new_variables([], InMap, [], InMap).
map_vars_to_new_variables([(V, out)|T], InMap, [VOut|TOut], OutMap) :-
    associate_var(V, InMap, VOut, OutMap1), !,
    map_vars_to_new_variables(T, OutMap1, TOut, OutMap).
map_vars_to_new_variables([(V, in)|T], InMap, [V|TOut], OutMap) :-
    map_vars_to_new_variables(T, InMap, TOut, OutMap).

associate_var('$VAR'(VarNum), InMap, '$VAR'(N), OutMap) :-
    get_assoc(VarNum, InMap, NMappings), !,
```

```prolog
322        b_getval(variable_index_counter, N),
323        NewNMappings = [N|NMappings],
324        N1 is N + 1,
325        b_setval(variable_index_counter, N1),
326        put_assoc(VarNum, InMap, NewNMappings, OutMap).
327  associate_var('$VAR'(VarNum), InMap, '$VAR'(VarNum), OutMap) :-
328        NMappings = [],
329        put_assoc(VarNum, InMap, NMappings, OutMap).
330
331  duplicate_literals([], _, _).
332  duplicate_literals([T1 | Tail], InMap, L) :-
333        !,
334        duplicate_literals(T1, InMap, T2),
335        duplicate_literals(Tail, InMap, Tail2),
336        append(T2, Tail2, L).
337  duplicate_literals(Term, InMap, OutLits) :-
338        zip_vars_with_direction(Term, Pred, ZippedVars),
339        duplicate_literal(Pred, [], ZippedVars, InMap, OutLits).
340
341  duplicate_literal(Pred, StartVars, [], _, OutLits) :-
342        reverse(StartVars, RevStartVars),
343        OutLit =.. [Pred | RevStartVars],
344        OutLits = [OutLit].
345  duplicate_literal(Pred, StartVars, [('$VAR'(V), in) | Tail], InMap,
       ↪ OutLits) :-
346        get_assoc(V, InMap, NMappings), !,
347        maplist(duplicate_literal_map(Pred, StartVars, Tail, InMap), [V |
           ↪ NMappings], NewLitsL),
348        flatten(NewLitsL, OutLits).
349  duplicate_literal(Pred, StartVars, [(V, _) | Tail], InMap, OutLits)
       ↪ :-
350        NewStartVars = [V | StartVars],
351        duplicate_literal(Pred, NewStartVars, Tail, InMap, OutLits).
352
353  duplicate_literal_map(Pred, StartVars, Tail, InMap, V, OutLits) :-
354        NewStartVars = ['$VAR'(V) | StartVars],
355        duplicate_literal(Pred, NewStartVars, Tail, InMap, OutLits).
356
357  %% Formatting for python
358  neg_bottom_plus_chains_for_python(X, S, C) :-
359        neg_bottom(X, B),
360        format_bottom_plus_chains_for_python(B, S, C).
361
362  pos_bottom_plus_chains_for_python(X, S, C) :-
363        pos_bottom(X, B),
364        format_bottom_plus_chains_for_python(B, S, C).
365
366  pos_bottom_complete_plus_chains_for_python(X, S, C, E) :-
367        pos_bottom(X, B),
368        complete_bottom_clause(B, BottomUnique, E1),
369        term_string(E1, E),
370        format_bottom_plus_chains_for_python(BottomUnique, S, C).
371
372  neg_bottom_complete_plus_chains_for_python(X, S, C, E) :-
```

```prolog
373      neg_bottom(X, B),
374      complete_bottom_clause(B, BottomUnique, E1),
375      term_string(E1, E),
376      format_bottom_plus_chains_for_python(BottomUnique, S, C).
377
378 format_bottom_plus_chains_for_python(B, S, C) :-
379      find_chains(B, Chains),
380      format_chains_for_python(Chains, ChainPairs),
381      format_clause_for_python(B, Term),
382      term_string(Term, S),
383      term_string(ChainPairs, C).
384
385 format_chains_for_python(Chains, ChainPairs) :-
386      assoc_to_list(Chains, ChainList),
387      maplist(pair_to_tuple, ChainList, ChainPairs).
388
389 pair_to_tuple(X-Y, (X, Y)).
390
391 format_clause_for_python(Head :- Body, Output) :-
392      list_from_tuple(Body, BodyL),
393      format_literal_for_python(Head, FormattedHead),
394      maplist(format_literal_for_python, BodyL, FormattedBody),
395      Output = (FormattedHead, FormattedBody), !.
396
397 format_literal_for_python(Literal, FormattedLiteral) :-
398      Literal =.. [Predicate | Arguments],
399      atom_string(Predicate, PredicateName),
400      maplist(format_argument_for_python, Arguments, ArgumentNames),
401      FormattedLiteral = (PredicateName, ArgumentNames).
402
403 format_argument_for_python('$VAR'(VarNumber), VarNumber).
404
405 % Just leaving Vars as numbers for now.
406 % format_argument_for_python('$VAR'(VarNumber), VarName) :-
407 %     char_code('A', AIndex),
408 %     divmod(VarNumber, 26, Quotient, Remainder),
409 %     Index is Remainder + AIndex,
410 %     char_code(C, Index),
411 %     (Quotient is 0 -> VarName = C;
412 %     atomic_concat(C, Quotient, VarName)).
413
414 format_argument_for_python(Argument, ArgumentFormatted) :-
415      atom(Argument), !,
416      atom_string(Argument, ArgumentFormatted).
417
418 % Not including this for now so that we get an error if we have
    ↪ compound arguments.
419 %format_argument_for_python(Argument, ArgumentFormatted) :-
420 %     compound(Argument), format_literal_for_python(Argument,
    ↪ ArgumentFormatted).
```

# Appendix C

# `bc_enc` code

The following file is entirely the author's own work. Unfortunately, it is dependent on a larger set of code from the full Popper implementation. The current version of Popper can be found here: https://github.com/logic-and-learning-lab/Popper

Other files were not included as they are not entirely the author's contribution.

```python
import clingo
from pyswip import Prolog
from ast import literal_eval

from .generate import var_name
from .tester import init_prolog as init_popper_prolog
from .core import Literal, Clause, Constraint

import itertools

NEGATIVE = "neg"
POSITIVE = "pos"
BASE_GROUNDING_PROGRAM = """
#show v_var/2.
v_val(0..num_v_vals-1).
:- v_var(X, _), not sat(X).
"""

## Format functions for bias predicates
def basic_pred_format(symbol):
    return f'{symbol.name}({symbol.arguments[0].name}, {str(symbol.arguments[1])})'

def other_pred_format(symbol):
    args = ', '.join([argument.name for argument in symbol.arguments[1].arguments])
    return f'{symbol.name}({symbol.arguments[0].name}, ({args}))'

BIAS_PREDS = [
        ("body_pred", 2, basic_pred_format),
        ("head_pred", 2, basic_pred_format),
        ("direction", 2, other_pred_format),
```

```python
            ("type", 2, other_pred_format),
        ]

class Preprocessor:
    def __init__(self, experiment):
        self.process_experiment_settings(experiment)

        self.init_solver()

        self.init_prolog()

    def process_experiment_settings(self, experiment):
        self.experiment = experiment

        self.complete = experiment.preprocessing.complete
        self.example_types = experiment.preprocessing.example_types
        self.debug = experiment.preprocessing.debug
        self.single_variant_positive =
            experiment.preprocessing.single_variant_positive
        self.max_variable_depth =
            experiment.preprocessing.max_variable_depth

    def init_solver(self):
        self.solver = clingo.Control(self.experiment.clingo_args)

        self.solver.add('bias', [], self.experiment.bias_string)

        self.solver.ground([('bias', [])])

        max_vars_atoms =
            self.solver.symbolic_atoms.by_signature('max_vars',
            arity=1)
        self.max_vars =
            next(max_vars_atoms).symbol.arguments[0].number

        if not self.max_variable_depth:
            self.max_variable_depth = self.max_vars - 1

    def init_prolog(self):
        self.prolog = Prolog()

        init_popper_prolog(self.prolog, [self.experiment.bk_filename,
            self.experiment.exs_filename], 'preprocess.pl')

        for (pred_name, pred_arity, pred_format_function) in
            BIAS_PREDS:
            pred_atoms =
                self.solver.symbolic_atoms.by_signature(pred_name,
                arity=pred_arity)
            for pred_symbol in pred_atoms:

                    self.prolog.assertz(pred_format_function(pred_symbol.symbol))

```

```python
                  ↪   self.prolog.assertz(f'max_variable_depth({self.max_variable_depth})')

    def preprocess_constraints(self):
        for example_type in self.example_types:
            yield from
                ↪   self.preprocess_constraints_by_type(example_type)

        if self.experiment.show_stats:
            self.experiment.preprocessing.print_stats()

    def preprocess_constraints_by_type(self, example_type):
        if self.complete:
            bottom_results =
                ↪   self.prolog.query(f"{example_type}_bottom_complete_plus_chains_fo
                ↪   B, C, E)")
        else:
            bottom_results =
                ↪   self.prolog.query(f"{example_type}_bottom_plus_chains_for_python(
                ↪   B, C)")
        for (i, bottom_result) in enumerate(bottom_results):
            clause = self.parse_clause(bottom_result["B"])
            chains = self.parse_dict(bottom_result["C"])
            equalities = self.parse_dict(bottom_result["E"]) if
                ↪   self.complete else {}

            clause_handle = f"{example_type}_bc_{i}"

            bc_constraint =
                ↪   Constraint(f"{example_type}_bottom_clause",
                ↪   Literal(f"{example_type}_bottom_clause",
                ↪   (clause_handle,)), [])

                ↪   self.experiment.preprocessing.register_bottom_clause(example_type
                ↪   clause, bc_constraint, chains, equalities)

            yield bc_constraint

            variants = self.find_variants(clause, chains, equalities,
                ↪   example_type)

            for (j, variant) in enumerate(variants):
                variant_handle = f"{clause_handle}_v_{j}"
                constraints = [Constraint("bottom_clause_variant",
                    ↪   Literal("bottom_clause_variant", (clause_handle,
                    ↪   variant_handle)), [])]
                constraints +=
                    ↪   self.constraints_for_variant(variant_handle,
                    ↪   clause, variant)

                    ↪   self.experiment.preprocessing.register_variant(clause,
                    ↪   variant, constraints)

```

```python
                yield from constraints

    def find_variants(self, clause, chains, equalities,
        example_type):
        program = self.generate_grounding_program(clause, chains,
            equalities, example_type)

        grounder = clingo.Control(['--rand-freq=0'])

        # ask for all models
        grounder.configuration.solve.models = 0

        grounder.add('base', [], program)

        grounder.ground([("base", [])])

        out = []

        # TODO Maybe clean this up. This is modified copied code from
            aspsolver.py
        # Can we reuse the grounder?
        def on_model(m):
            xs = m.symbols(shown = True)
            # map a variable to a program variable
            assignment = {}
            for x in xs:
                var = x.arguments[0].number
                val = x.arguments[1].number
                assignment[var] = val
            out.append(assignment)

        grounder.solve(on_model=on_model)
        return out

    def generate_grounding_program(self, clause, chains, equalities,
        example_type):
        program = self.initialize_grounding_program(clause, chains,
            equalities)

        if example_type == POSITIVE and self.single_variant_positive:
            program.extend(self.positive_grounding_constraints())
        else:
            program.extend(self.variant_grounding_constraints(clause,
                chains))

        return "\n".join(program)

    def initialize_grounding_program(self, clause, chains,
        equalities):
        program = [BASE_GROUNDING_PROGRAM]

        program.append(f"#const num_v_vars={len(clause.all_vars)}.")
        program.append(f"#const num_v_vals={self.max_vars}.")

```

```python
157            is_complete = self.complete and equalities
158            if is_complete:
159                program.append(":- v_val(X), v_var(I, X), v_var(J, X), I != J, not eq_var(I, J).")
160
161            if len(clause.all_vars) > self.max_vars:
162                program.append("0 {v_var(V,X): v_val(X)} 1:- V=0..num_v_vars-1.")
163                if is_complete:
164                    program.append(":- v_val(X), #count{I : v_var(I,X)} < 1.")
165                else:
166                    program.append(":- v_val(X), #count{I : v_var(I,X)} != 1.")
167            else:
168                if is_complete:
169                    program.append("1 {v_var(V,X): v_val(X)}:- V=0..num_v_vars-1.")
170                else:
171                    program.append("1 {v_var(V,X): v_val(X)} 1:- V=0..num_v_vars-1.")
172                    program.append(":- v_val(X), #count{I : v_var(I,X)} > 1.")
173
174            head_args = set(clause.head.arguments)
175            for v in head_args:
176                program.append(f"v_var({v}, {v}).")
177                program.append(f"sat({v}).")
178
179            for v in clause.all_vars:
180                if v not in chains:
181                    if v not in head_args:
182                        program.append(f"sat({v}).")
183                    continue
184
185                dependencies = chains[v]
186                for dependency in dependencies:
187                    d_vars = list(dependency) if isinstance(dependency, tuple) else [dependency]
188                    sats = [f"sat({d_var}), v_var({d_var}, _)" for d_var in d_vars]
189                    program.append(f"sat({v}) :- {', '.join(sats)}.")
190
191            for (k, v) in equalities.items():
192                var_set = set([k] + v)
193                for (x, y) in itertools.combinations(var_set, 2):
194                    program.append(f"eq_var({x}, {y}).")
195                    program.append(f"eq_var({y}, {x}).")
196
197            return program
198
199    def positive_grounding_constraints(self):
200
```

```python
201            # Variables must be assigned indexes in ascending order. This
               # enforces a single syntactic variant.
202            return [":- v_var(I, X), v_var(J, Y), X < Y, I > J."]

204    def variant_grounding_constraints(self, clause, chains):

206        head_args = set(clause.head.arguments)
207        depths = {}
208        for arg in head_args:
209            depths[arg] = 0

211        depths = self.compute_depths(clause, chains, depths)

213        depth_offset_for_head = len(head_args) - 1

215        constraints = []
216        for (v, depth) in depths.items():
217            if v in head_args:
218                continue
219            constraints.append(f"depth({v}, {depth + 
                depth_offset_for_head}).")

221        # Variables must have indices equal to or higher than their
        # depth.
222        constraints.append(":- v_var(X, I), depth(X, J), I < J.")

224        return constraints

226    def constraints_for_variant(self, variant_handle, clause,
        variant):
227        yield self.bottom_clause_head_constraint(variant_handle,
            clause, variant)

229        for body_lit in clause.body:
230            if not all(arg in variant for arg in body_lit.arguments):
231                continue

233            swapped_args = tuple([variant[arg] for arg in
                body_lit.arguments])
234            yield self.bottom_clause_constraint("bottom_clause_body",
                variant_handle, body_lit, swapped_args)

236    def bottom_clause_head_constraint(self, variant_handle, clause,
        variant):
237        # TODO(Brad): This is just a sanity check. I should probably
            # remove it or make it better after testing.
238        for arg in clause.head.arguments:
239            if variant[arg] != arg:
240                raise Exception("Something went wrong! Head variables
                    were not assigned their index.")

242        return self.bottom_clause_constraint("bottom_clause_head",
            variant_handle, clause.head, clause.head.arguments)

```

```python
244        def bottom_clause_constraint(self, predicate, variant, literal,
           ↪ arguments):
245            args = tuple([variant, literal.predicate, arguments])
246            return Constraint(predicate, Literal(predicate, args), [])
247
248        def parse_clause(self, clause_string):
249            # This assumes we always get a tuple
250            ((head_pred, head_args), body_literals) =
               ↪ literal_eval(clause_string.decode('UTF-8').strip())
251            head = Literal(head_pred, tuple(head_args))
252            body = [Literal(body_pred, tuple(body_args)) for (body_pred,
               ↪ body_args) in body_literals]
253            return Clause(head, body)
254
255        def parse_dict(self, dict_string):
256            return
               ↪ dict(literal_eval(dict_string.decode('UTF-8').strip()))
257
258        def compute_depths(self, clause, chains, depths):
259            for variable_name in sorted(clause.all_vars):
260                self.compute_depth(variable_name, chains, depths, set())
261
262            return depths
263
264        def compute_depth(self, variable_name, chains_dict, depths_dict,
           ↪ path):
265            if variable_name in depths_dict:
266                return depths_dict[variable_name]
267
268            if variable_name not in chains_dict:
269                depths_dict[variable_name] = 1
270                return 1
271
272            if variable_name in path:
273                return None
274
275            min_depth = None
276            for dependency in chains_dict[variable_name]:
277                if isinstance(dependency, tuple):
278                    results = [self.compute_depth(dependency_name,
                       ↪ chains_dict, depths_dict,
                       ↪ path.union({variable_name})) for dependency_name
                       ↪ in dependency]
279                    filtered = [v for v in results if v != None]
280                    depth = min(filtered) if filtered else None
281                else:
282                    depth = self.compute_depth(dependency, chains_dict,
                       ↪ depths_dict, path.union({variable_name}))
283                if min_depth == None:
284                    min_depth = depth
285                elif depth != None and depth < min_depth:
286                    min_depth = depth
287            if min_depth != None:
288                depths_dict[variable_name] = min_depth + 1
```

```python
289                return min_depth + 1
290            else:
291                return None
```

# Appendix D

# `hyp_constraints` and `gen_rels` code

```prolog
%% PREPROCESSING CONSTRAINTS
does_not_generalize_bottom_clause_variant(Cl, V):-
    bottom_clause_variant(_, V),
    head_literal(Cl, P, _, Vars),
    not bottom_clause_head(V, P, Vars).
does_not_generalize_bottom_clause_variant(Cl, V):-
    bottom_clause_variant(_, V),
    body_literal(Cl, P, _, Vars),
    not bottom_clause_body(V, P, Vars).

does_not_recursively_generalize_bottom_clause_variant(Cl, V):-
    bottom_clause_variant(_, V),
    precedes_recursion(Cl, P, Vars),
    not bottom_clause_body(V, P, Vars).

generalizes_bottom_clause(Cl, E) :-
    clause(Cl),
    bottom_clause_variant(E, V),
    not does_not_generalize_bottom_clause_variant(Cl, V).

generalizes_bottom_clause(Cl, E) :-
    recursive,
    clause(Cl),
    pos_bottom_clause(E),
    bottom_clause_variant(E, V),
    not does_not_recursively_generalize_bottom_clause_variant(Cl, V).

:-
    not enable_pi,
    neg_bottom_clause(E),
    generalizes_bottom_clause(_, E).

:-
    not enable_pi,
    not relaxed_positive_bottom_clause_constraints,
    pos_bottom_clause(E),
```

```prolog
37        not generalizes_bottom_clause(_, E).

38
39  generalizes_positive_bottom_clause(Cl) :-
40        pos_bottom_clause(E),
41        generalizes_bottom_clause(Cl, E).

42
43  :-
44        not enable_pi,
45        relaxed_positive_bottom_clause_constraints,
46        clause(Cl),
47        pos_bottom_clause(_),
48        not generalizes_positive_bottom_clause(Cl).

49
50  % Recursion handling for predicate invention
51  var_direction(P, Vars, Dir, Var) :-
52        direction_(P, Pos, Dir),
53        var_pos(Var, Vars, Pos).

54
55  in_var_is_out_var(Cl, PI, VarsI, P, Vars) :-
56        var_direction(PI, VarsI, in, Var),
57        body_literal(Cl, P, _, Vars),
58        var_direction(P, Vars, out, Var).

59
60  precedes_recursion(Cl, P, Vars):-
61        head_literal(Cl, PI, AI, _),
62        body_literal(Cl, PI, AI, VarsI),
63        in_var_is_out_var(Cl, PI, VarsI, P, Vars).
64  precedes_recursion(Cl, P, Vars):-
65        precedes_recursion(Cl, PI, VarsI),
66        in_var_is_out_var(Cl, PI, VarsI, P, Vars).
```

# Appendix E

# List manipulation problems files

## E.1 Bias

```
1  max_vars(5).
2  max_body(5).
3  max_clauses(3).
4
5  enable_recursion.
6
7  body_pred(head,2).
8  body_pred(tail,2).
9  body_pred(geq,2).
10 body_pred(empty,1).
11 body_pred(even,1).
12 body_pred(odd,1).
13 body_pred(one,1).
14 body_pred(zero,1).
15 body_pred(decrement,2).
16
17 type(cons,(element,list,list)).
18 direction(cons,(in,in,out)).
19
20 type(head,(list,element)).
21 direction(head,(in,out)).
22
23 type(tail,(list,list)).
24 direction(tail,(in,out)).
25
26 type(empty,(list,)).
27 direction(empty,(in,)).
28
29 type(element,(list,element)).
30 direction(element,(in,out)).
31
32 type(increment,(element,element)).
33 direction(increment,(in,out)).
34
35 type(decrement,(element,element)).
36 direction(decrement,(in,out)).
```

```
37
38  type(geq,(element,element)).
39  direction(geq,(in,in)).
40
41  type(even,(element,)).
42  direction(even,(in,)).
43
44  type(odd,(element,)).
45  direction(odd,(in,)).
46
47  type(one,(element,)).
48  direction(one,(in,)).
49
50  type(zero,(element,)).
51  direction(zero,(out,)).
52
53  type(sum,(element,element,element)).
54  direction(sum,(in,in,out)).
```

### E.1.1  dropk

```
1  head_pred(f,3).
2  type(f,(list,element,list)).
3  direction(f,(in,in,out)).
```

### E.1.2  droplast

```
1  head_pred(f,2).
2  type(f,(list,list)).
3  direction(f,(in,out)).
4
5  body_pred(cons,3).
```

### E.1.3  evens

```
1  head_pred(f,1).
2  type(f,(list,)).
3  direction(f,(in,)).
```

### E.1.4  finddup

```
1  head_pred(f,2).
2  type(f,(list, element)).
3  direction(f,(in,out)).
4
5  %% ESSENTIAL
6  body_pred(element,2).
```

### E.1.5 last

```
1  head_pred(f,2).
2  type(f,(list, element)).
3  direction(f,(in,out)).
```

### E.1.6 len

```
1  head_pred(f,2).
2  type(f,(list,element)).
3  direction(f,(in,out)).
4
5  body_pred(increment,2).
```

### E.1.7 member

```
1  head_pred(f,2).
2  type(f,(list,element)).
3  direction(f,(in,out)).
```

### E.1.8 sorted

```
1  head_pred(f,1).
2  type(f,(list,)).
3  direction(f,(in,)).
```

### E.1.9 sumlist

```
1  head_pred(f,2).
2  type(f,(list, element)).
3  direction(f,(in,out)).
4
5  body_pred(sum,3).
```

## E.2 BK

```
1  is_list([]).
2  is_list([_|_]).
3
4  increment(A,B):-
5      (nonvar(A) -> \+ is_list(A); true),
6      (nonvar(B) -> \+ is_list(B); true),
7      succ(A,B).
8
9  decrement(A,B):-
10     (nonvar(A) -> \+ is_list(A); true),
11     (nonvar(B) -> \+ is_list(B); true),
12     succ(B,A).
13
14 my_length(A,B):-
15     (nonvar(A) -> is_list(A); true),
```

```prolog
16      (nonvar(B) -> \+is_list(B); true),
17      length(A,B).

19  cons(A,B,C):-
20      append([A],B,C).
21  comps([H|T],H,T).

23  tail([_|T],T).
24  head([H|_],H).
25  sum(A,B,C):-
26      (nonvar(A) -> \+ is_list(A); true),
27      (nonvar(B) -> \+ is_list(B); true),
28      (nonvar(C) -> \+ is_list(B); true),
29      C is A+B.
30  empty([]).

32  element([X|_],X):-!.
33  element([_|T],X):-
34      element(T,X).

36  empty_in([]).
37  empty_out([]).

39  zero(0).
40  one(1).

42  gt(A,B):-
43      nonvar(A),
44      nonvar(B),
45      \+is_list(A),
46      \+is_list(B),
47      A > B.

49  geq(A,B):-
50      nonvar(A),
51      nonvar(B),
52      \+is_list(A),
53      \+is_list(B),
54      A >= B.

56  even(A):-
57      nonvar(A),
58      \+ is_list(A),
59      0 is A mod 2.

61  odd(A):-
62      nonvar(A),
63      \+ is_list(A),
64      1 is A mod 2.
```

# Appendix F

# Trains problem files

## F.1 Bias

```
1  max_clauses(4).
2  max_vars(5).
3  max_body(5).
4
5  head_pred(f,1).
6  body_pred(has_car,2).
7  body_pred(has_load,2).
8  body_pred(long,1).
9  body_pred(short,1).
10 body_pred(two_wheels,1).
11 body_pred(three_wheels,1).
12 body_pred(roof_arc,1).
13 body_pred(roof_closed,1).
14 body_pred(roof_flat,1).
15 body_pred(roof_jagged,1).
16 body_pred(roof_open,1).
17 body_pred(roof_peaked,1).
18 body_pred(zero_load,1).
19 body_pred(one_load,1).
20 body_pred(two_load,1).
21 body_pred(three_load,1).
22 body_pred(circle,1).
23 body_pred(diamond,1).
24 body_pred(hexagon,1).
25 body_pred(inverted_triangle,1).
26 body_pred(rectangle,1).
27 body_pred(triangle,1).
28
29 type(f,(train,)).
30 type(has_car,(train,car)).
31 type(has_load,(car,load)).
32 type(long,(car,)).
33 type(short,(car,)).
34 type(two_wheels,(car,)).
35 type(three_wheels,(car,)).
36 type(roof_arc,(car,)).
```

```
37  type(roof_closed,(car,)).
38  type(roof_flat,(car,)).
39  type(roof_jagged,(car,)).
40  type(roof_open,(car,)).
41  type(roof_peaked,(car,)).
42  type(zero_load,(load,)).
43  type(one_load,(load,)).
44  type(two_load,(load,)).
45  type(three_load,(car,)).
46  type(circle,(load,)).
47  type(diamond,(load,)).
48  type(hexagon,(load,)).
49  type(inverted_triangle,(load,)).
50  type(rectangle,(load,)).
51  type(triangle,(load,)).
52
53
54  direction(f,(in,)).
55  direction(has_car,(in,out)).
56  direction(has_load,(in,out)).
57  direction(long,(in,)).
58  direction(short,(in,)).
59  direction(two_wheels,(in,)).
60  direction(three_wheels,(in,)).
61  direction(roof_arc,(in,)).
62  direction(roof_closed,(in,)).
63  direction(roof_flat,(in,)).
64  direction(roof_jagged,(in,)).
65  direction(roof_open,(in,)).
66  direction(roof_peaked,(in,)).
67  direction(zero_load,(in,)).
68  direction(one_load,(in,)).
69  direction(two_load,(in,)).
70  direction(three_load,(in,)).
71  direction(circle,(in,)).
72  direction(diamond,(in,)).
73  direction(hexagon,(in,)).
74  direction(inverted_triangle,(in,)).
75  direction(rectangle,(in,)).
76  direction(triangle,(in,)).
```

## F.2 BK

The BK for the trains problem is very large. It can be found here: https://github.com/logic-and-learning-lab/Popper/blob/main/examples/trains/bk.pl